\documentclass[11pt, letterpaper, logo, onecolumn, copyright]{main}

\usepackage[numbers, sort&compress]{natbib}

\usepackage[inkscapeformat=png]{svg}

\usepackage[most, breakable, skins]{tcolorbox}

\tcbuselibrary{skins}
\usepackage{lipsum}
\usepackage{tabularx}
\usepackage{afterpage}
\usepackage{booktabs}
\usepackage{subcaption}
\usepackage{makecell}
\usepackage{multirow}
\usepackage{bm}
\usepackage[dvipsnames]{xcolor}
\usepackage{multicol}
\usepackage{array}
\usepackage{float}
\usepackage{listings, listings-rust}
\usepackage{fontawesome5}
\usepackage{hyperref}
\usepackage{amssymb,graphicx}
\usepackage{cleveref}
\usepackage{longtable}
\usepackage{pdflscape}
\usepackage{adjustbox}
\usepackage{nicematrix}
\usepackage{CJKutf8}
\usepackage{ragged2e}
\usepackage{colortbl}
\usepackage{enumitem}
\usepackage[ruled,linesnumbered]{algorithm2e}
\usepackage{pifont}
\usepackage[htt]{hyphenat}
\usepackage{amsmath}
\usepackage{amsthm}  
\usepackage{mathrsfs} 
\usepackage{circledsteps}
\usepackage{diagbox}
\usepackage{bookmark}

\usepackage{wrapfig}
\usepackage{xspace}
\usepackage{tikz}
\usepackage[normalem]{ulem}

\usepackage{pgfplots}
\usepackage{pgfplotstable}
\pgfplotsset{compat=1.18}

\usepackage{graphicx}
\usepackage{amsmath}
\usepackage{subcaption}
\usepackage{titletoc}
\usepackage{marvosym}
\usepackage{xstring}
\usepackage{multirow}
\usepackage[table,xcdraw]{xcolor}
\usepackage[normalem]{ulem}
\useunder{\uline}{\ul}{}
\usepackage{pgfplots}
\pgfplotsset{compat=1.18}
\usepackage{booktabs}
\usepackage{graphicx}
\usepackage{caption}
\usepackage{subcaption}
\usepackage{tikz}
\usetikzlibrary{arrows.meta}
\usepackage{colortbl}

\definecolor{medgray55}{gray}{0.55}
\definecolor{medgray}{gray}{0.7}
\definecolor{litegray}{gray}{0.9}
\definecolor{gblue}{RGB}{210, 227, 252}
\definecolor{gred}{RGB}{250, 210, 207}
\definecolor{gyellow}{RGB}{254, 239, 195}
\definecolor{ggreen}{RGB}{206, 234, 214}
\definecolor{gorange}{RGB}{254, 223, 200}

\definecolor{gblue9}{RGB}{23, 78, 166}
\definecolor{gred9}{RGB}{165, 14, 14}
\definecolor{gyellow9}{RGB}{227, 116, 0}
\definecolor{ggreen9}{RGB}{13, 101, 45}
\definecolor{gorange9}{RGB}{176, 96, 0}

\definecolor{myblue}{rgb}{0,0,1}
\definecolor{myred}{rgb}{1,0,0}
\definecolor{mylightgray}{gray}{0.95}
\definecolor{myCite}{HTML}{1C4587}

\definecolor{highlightblue}{HTML}{185ABC}
\definecolor{cellHighlight}{HTML}{dbefff}

\usepackage{minitoc}

\noptcrule


\newcolumntype{L}[1]{>{\raggedright\let\newline\\\arraybackslash\hspace{0pt}}m{#1}}
\newcolumntype{C}[1]{>{\centering}m{#1}}

\newcolumntype{R}[1]{>{\raggedleft\let\newline\\\arraybackslash\hspace{0pt}}m{#1}}

\definecolor{ao}{rgb}{0.0, 0.0, 1.0}

\newcommand\vcent[1]{\vcenter{\hbox{#1}}}

\newcommand\loudspeaker[1][3]{\ensuremath{\vcent{\rule{.6ex}{.6ex}}\kern-.5ex
  \vcent{\scalebox{.6}[1]{\rotatebox[origin=center]{90}{$\blacktriangle$}}}
  \ifnum#1>0\relax\kern.05ex\vcent{\scalebox{.4}{\ttfamily)}}
  \ifnum#1>1\relax\kern-.4ex\vcent{\scalebox{.56}{\ttfamily)}}
  \ifnum#1>2\relax\kern-.55ex\vcent{\scalebox{.7}{\ttfamily)}}
  \fi\fi\fi}
}

\makeatletter
\renewcommand\subparagraph{
 \@startsection {subparagraph}{5}{\z@ }{3.25ex \@plus 1ex
 \@minus .2ex}{-1em}{\normalfont \normalsize \bfseries }}
\makeatother

\let\cite\citep
\hypersetup{
  citecolor = myCite,  
  linkcolor = myCite,   
  urlcolor  = myCite
}

\title{FRoM-W1: Towards General Humanoid Whole-Body Control with Language Instructions}

\author{
    Peng Li$^{1,2}$$^{* \dag}$ Zihan Zhuang$^1$$^{* \dag}$ Yangfan Gao$^1$$^*$ Yi Dong$^1$$^*$ Sixian Li$^{1,2}$ Changhao Jiang$^1$\\
\textbf{Shihan Dou$^1$ Zhiheng Xi$^1$ Enyu Zhou$^1$ Jixuan Huang$^1$ Hui Li$^1$ JingJing Gong$^{2}$ Xingjun Ma$^{1,2}$} \\
\textbf{Tao Gui$^{1}$\textsuperscript{\Letter} Zuxuan Wu$^{1,2}$ Qi Zhang$^1$ Xuanjing Huang$^1$ Yu-Gang Jiang$^1$ Xipeng Qiu$^{1,2}$\textsuperscript{\Letter}}
\\
$^1$Fudan University $^2$Shanghai Innovation Institute \\
\texttt{pengli@sii.edu.cn, \{zhzhuang24, gaoyf24, yidong25, sxli25, chjiang25\}@m.fudan.edu.cn\\  \{tgui, xpqiu\}@fudan.edu.cn} \\
Fully Open-Sourced at \href{https://openmoss.github.io/FRoM-W1}{https://openmoss.github.io/FRoM-W1}
}
\vspace{-50pt}

\newpage
\begin{abstract} 

Humanoid robots are capable of performing various actions such as greeting, dancing and even backflipping. However, these motions are often hard-coded or specifically trained, which limits their versatility.
In this work, we present \texttt{FRoM-W1}, an open-source framework designed to achieve general humanoid whole-body motion control using natural language.
To universally understand natural language and generate corresponding motions, as well as enable various humanoid robots to stably execute these motions in the physical world under gravity, \texttt{FRoM-W1} operates in two stages:
(a) \texttt{H-GPT}: utilizing massive human data, a large-scale language-driven human whole-body motion generation model is trained to generate diverse natural behaviors. 
We further leverage the Chain-of-Thought technique to improve the model's generalization in instruction understanding.
(b) \texttt{H-ACT}: After retargeting generated human whole-body motions into robot-specific actions, a motion controller that is pretrained and further fine-tuned through reinforcement learning in physical simulation enables humanoid robots to accurately and stably perform corresponding actions. 
It is then deployed on real robots via a modular simulation-to-reality module.
We extensively evaluate \texttt{FRoM-W1} on Unitree H1 and G1 robots. 
Results demonstrate superior performance on the HumanML3D-X benchmark for human whole-body motion generation, and our introduced reinforcement learning fine-tuning consistently improves both motion tracking accuracy and task success rates of these humanoid robots.
We open-source the entire \texttt{FRoM-W1} framework and hope it will advance the development of humanoid intelligence.
\end{abstract}

\begin{document}

\doparttoc
\faketableofcontents

\begingroup
  \renewcommand\thefootnote{}
  \footnote{\textsuperscript{*}Equal contribution.
            \textsuperscript{\dag}Project Lead.
            \textsuperscript{\Letter}Corresponding authors.
            }
    % \footnote{\textsuperscript{1}Our code are available at \url{https://github.com/XXXXXXXXXX}.}
  \addtocounter{footnote}{-1}
\endgroup

\vspace{-30pt}
\maketitle

\section{Introduction}
Humanoid robots, such as Unitree’s H1 and G1, feature an anthropomorphic design that mimics the human form, enabling them to operate and interact seamlessly in environments around us~\citep{DBLP:journals/corr/abs-2501-02116, DBLP:journals/access/KawaharazukaOYPZ25}.
Current technologies allow these robots to perform impressive motions—including greeting gestures, dancing, practicing kung fu, and even executing backflip. However, such capabilities are typically achieved through hardcoded trajectories or task-specific reinforcement learning, often controlled via handheld devices~\citep{DBLP:conf/humanoids/ChignoliKSK21, DBLP:conf/corl/ZhuangYZ24, DBLP:journals/corr/abs-2410-11825, DBLP:journals/corr/abs-2502-10363, DBLP:journals/corr/abs-2506-12851, DBLP:journals/corr/abs-2508-21043}.
These approaches limit the humanoid robot's ability to autonomously perceive and interact with the external world, particularly with humans.
To enable natural human-robot interaction, this work addresses the following core research question: \textbf{How can humanoid robots comprehend diverse natural language instructions and execute corresponding whole-body motions in the real physical world?}

\begin{figure}[t]
    \centering
    \includegraphics[page=1, width=1\textwidth]{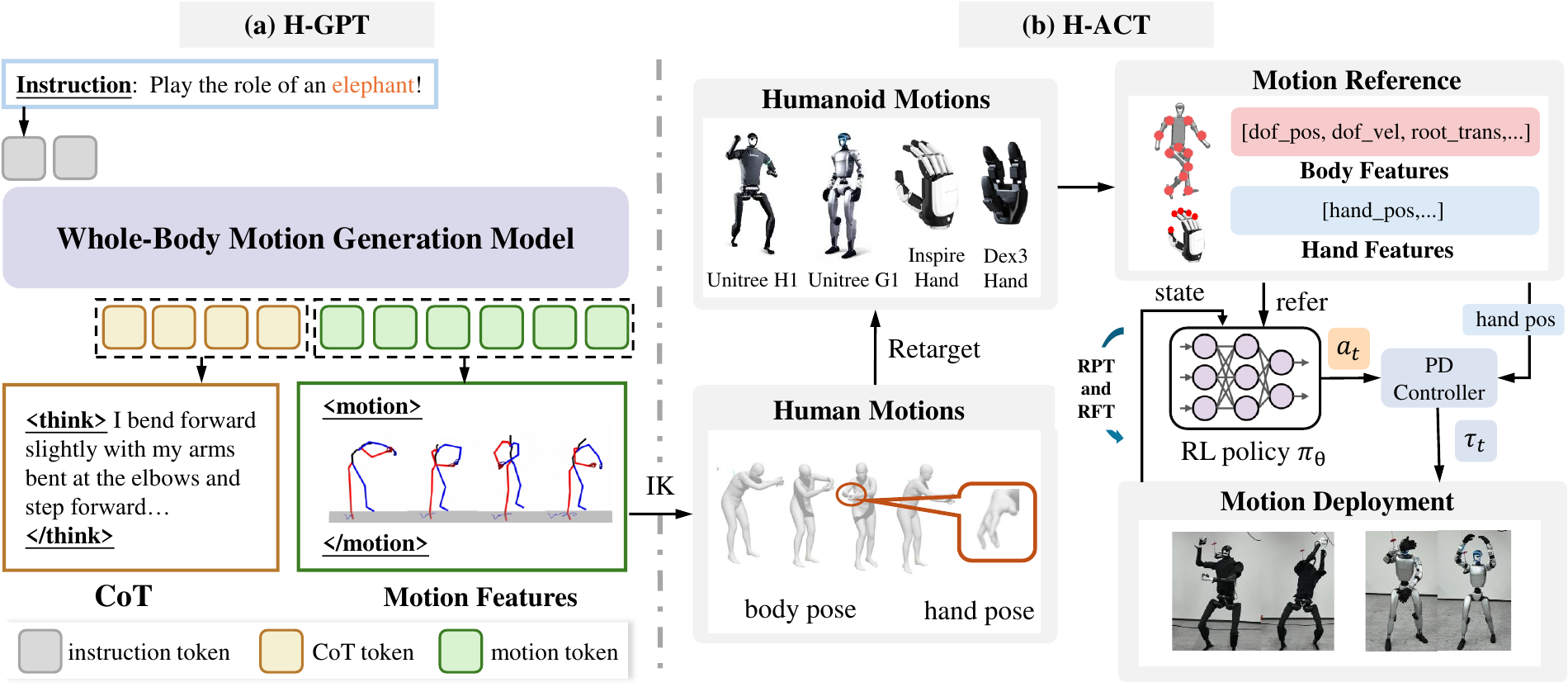} % trim=10 10 10 10, clip,
    \caption{
        The inference pipeline of 
        \texttt{FRoM-W1}.
        (a) \texttt{H-GPT} first translates language instructions into fine-grained action descriptions through CoT, then generates corresponding whole-body human motion sequences. 
        (b) After obtaining the human SMPL-X motions through inverse kinematics (IK), \texttt{H-ACT} retargets the generated human motion sequences into robot-specific motions. Then, through a reinforcement learning pre-trained (RPT) and fine-tuned (RFT) whole-body motion controller, \texttt{H-ACT} enables the humanoid robot to perform the corresponding actions in the physical world.
       % \textcolor{red}{IK;AR Tokens} Humanoid Motions;Better hand pose;
    }
    \label{fig:overview}
\end{figure}

Achieving this goal presents two major challenges.
First, like current large language models (LLMs) ~\citep{DBLP:conf/nips/Ouyang0JAWMZASR22, DBLP:journals/ijautcomp/SunZHLCLYSTZZCZZLZZLYWY24, DBLP:journals/corr/abs-2407-21783, DBLP:journals/corr/abs-2505-09388}, achieving general language understanding and motion generation typically requires vast data and training large models.
But there is a scarcity of large-scale paired robot datasets that associate language instructions with whole-body humanoid motions~\citep{DBLP:conf/iros/HeLXZKLS24, DBLP:journals/corr/abs-2506-08931, DBLP:journals/corr/abs-2510-08807}.
%
% Although some current efforts are attempting to collect such data through teleoperation~\citep{DBLP:conf/iros/HeLXZKLS24, DBLP:journals/corr/abs-2506-08931, DBLP:journals/corr/abs-2510-08807}, these methods require significant human labor and are difficult to scale.
% % \doney{why datasets is important? You need to mention foundation model first and then point out that FM requires datasets} 
Second, even if a foundation model were trained to generate robot motions from natural language, directly executing these motions on physical bipedal humanoid robots would likely lead to instability and falls due to gravity and dynamic uncertainties~\citep{DBLP:journals/corr/abs-2505-07294, meng2025safefalllearningprotectivecontrol}.

Inspired by the analogous roles of the human cerebrum and cerebellum in functions such as semantic comprehension and motion planning, as well as in stable motor execution~\citep{kandel2000principles, gazzaniga2006cognitive}, we propose the \texttt{FRoM-W1}\footnote{``\texttt{FRoM-W1}'' denotes \underline{F}oundational Humanoid \underline{Ro}bot \underline{M}odel -- \underline{W}hole-Body Control, Version \underline{1}''.} framework to address these challenges, as illustrated in Figure~\ref{fig:overview}.
First, although a large-scale annotated humanoid motion dataset is unavailable, the morphological similarity between humanoid robots and humans enables us to leverage extensive language-labeled human motion data from motion capture and videos~\citep{DBLP:conf/cvpr/GuoZZ0JL022, DBLP:conf/nips/LinZLCZWZ23, DBLP:journals/corr/abs-2507-07095}. 
Based on these resources, we employ the vector quantized variational autoencoder (VQ-VAE) technique~\citep{DBLP:conf/nips/OordVK17} as used in T2M-GPT~\citep{DBLP:journals/corr/abs-2301-06052} and MotionGPT~\citep{DBLP:conf/nips/JiangCLYYC23} to first convert human whole-body motion sequences, including hand movements, into token sequences aligned with language tokens.
% Using these resources, we first convert human whole-body motion sequences (including hands) into token sequences aligned withlanguage tokens, employing the vector quantized variational autoencoder (VQ-VAE) technique~\citep{DBLP:conf/nips/OordVK17} as used in T2M-GPT~\citep{DBLP:journals/corr/abs-2301-06052} and MotionGPT~\citep{DBLP:conf/nips/JiangCLYYC23}. 
%
Based on the LLaMA-3.1 LLM~\citep{DBLP:journals/corr/abs-2407-21783}, we train a generative model termed \texttt{H-GPT}, which synthesizes whole-body human motions from natural language instructions like \textit{``Play the role of an elephant!''}.
Chain-of-Thought (CoT)~\citep{DBLP:conf/nips/Wei0SBIXCLZ22, DBLP:journals/corr/abs-2501-12948} technology is used to decompose each instruction into body-level motion primitives with explicit temporal structure, enabling the model to convert abstract or complex instructions into simple, unified textual representations that facilitate motion generation.
Next, to enable humanoid robots of different morphologies to accurately and stably execute the generated human motions, we first retarget the human whole-body SMPL-X motions~\citep{DBLP:conf/cvpr/PavlakosCGBOTB19} to each specific robot platform.
Drawing inspiration from prior work on motion mimicking~\citep{DBLP:conf/iros/HeLXZKLS24, he2024omnih2ouniversaldexteroushumantohumanoid, DBLP:journals/corr/abs-2505-02833, DBLP:journals/corr/abs-2508-08241}, we then train a whole-body humanoid controller via reinforcement learning in the physical simulation Isaacgym~\cite{DBLP:conf/nips/MakoviychukWGLS21}.
In addition to large-scale pretraining of general motion controllers, we further introduce reinforcement learning fine-tuning (RFT) during the inference phase, enabling the controller to more accurately track the generated motions while maintaining whole-body stability.
Finally, we deploy the controller to various physical humanoid robots using a modular deployment interface.
We refer to these humanoid motion execution components collectively as \texttt{H-ACT}.

We conducted a comprehensive evaluation of the proposed framework using the Unitree H1 and G1 robots. 
On our constructed instruction to human whole-body motion generation benchmark HumanML3D-X, the framework achieved a 2.5-fold improvement in the primary metric Frechet Inception Distance (FID) compared to the baseline model T2M-GPT~\citep{DBLP:journals/corr/abs-2301-06052}. 
We have also developed a motion generation generalization evaluation benchmark called $\delta$HumanML3D-X and validated the effectiveness of incorporating the CoT method.
Moreover, we evaluated the effectiveness of our reinforcement learning fine-tuning strategy for the controller, which not only consistently increased the motion tracking success rate but also delivered a 15\% improvement in the accuracy metric Mean Per-joint Joint Position Error (MPJPE).
Furthermore, we demonstrate the versatility of our designed sim2real framework for humanoid robots. In addition to supporting our default control policy deployment, it also universally enables the efficient deployment of several recent motion mimicing controllers, such as HugWBC~\citep{xue2025unifiedgeneralhumanoidwholebody} and TWIST~\citep{DBLP:journals/corr/abs-2505-02833}.

Overall, the main contributions of \texttt{FRoM-W1}, a framewzork that enables whole-body control of humanoid robots using natural language instructions, are as follows:
\begin{itemize}
\item We introduce \texttt{H-GPT}, a 9B model that generates high quality whole-body human motions from natural language instructions. Enhanced with CoT technology, it achieves versatile instruction understanding. \texttt{H-GPT} demonstrates superior accuracy and generalization in human whole-body motion generation on our established HumanML3D-X and $\delta$HumanML3D-X benchmarks.
\item We propose \texttt{H-ACT}, a module that spans from human-to-humanoid motion retargeting to humanoid policy deployment, enabling humanoid robots to accurately and stably perform whole-body motions generated by \texttt{H-GPT}. The novel two-stage RL strategy of pretraining and fine-tuning allows the robot to more precisely track the corresponding motion.
\item We fully open-source the complete framework, including the \texttt{H-GPT} and \texttt{H-ACT} model training code, model checkpoints, evaluation benchmarks, and a lightweight deployment module, to facilitate further research and development in language-guided whole-body control for humanoid robots.
\end{itemize}

\section{\texttt{FRoM-W1}}
\label{sec:method}

In this section, We detail the design of the \texttt{FRoM-W1} framework.
We begin by introducing the task of whole-body motion control of humanoid robots using natural language instructions in Section \ref{sec:task_formulation}.
    Then, in Section \ref{sec:hgpt}, we elaborate on the design and training of the instruction to whole-body human motion generation model \texttt{H-GPT}. 
In Section \ref{sec:hact}, we describe \texttt{H-ACT}, covering its human-to-humanoid retargeting, reinforcement learning pre-training and fine-tuning, and the modular framework for deployment on real humanoid robots.

% % % % % % % % % % % % % % % % % % % % %  TASK FORMULATION % % % % % % % % % % % % % % % % % % % % % % % % % % % % % 

\subsection{Task Formulation}
\label{sec:task_formulation}

Our goal is to enable a physical humanoid robot to perform whole-body motions based on open-ended natural language instructions. The core task is to learn a mapping from a language instruction \( I \in \mathcal{I} \) like \textit{``Play the role of an elephant!''} to a sequence of executable and stable humanoid robot actions \( A_r \in \mathcal{A}_{r} \).
This task is decomposed into two sequential stages, corresponding to the \texttt{H-GPT} and \texttt{H-ACT} modules of our framework:

\textbf{Stage 1: Language to Whole-Body Human Motion (\texttt{H-GPT}).} Given an instruction \( I \), generate a semantically aligned and physically plausible whole-body human motion sequence \( M_{h} \in \mathcal{M}_{h} \). This stage learns the mapping $\mathcal{G}_{\text{H-GPT}}: \mathcal{I} \rightarrow \mathcal{M}_{h}$.

\textbf{Stage 2: Human Motion to Humanoid Execution (\texttt{H-ACT}).} Given the generated human motion \( M_{h} \), execute it stably on a target humanoid robot. This stage learns the mapping: $\mathcal{A}_{\text{H-ACT}}: \mathcal{M}_{h} \rightarrow \mathcal{A}_{r}$.
The function \( \mathcal{A}_{\text{H-ACT}} \) encompasses: 1) Kinematic retargeting \( \mathcal{R} \) to adapt the human motion \( M_{h}\) to the humanoid robot's morphology, producing \( M_{r} \in \mathcal{M}_r \), i.e., \( R: \mathcal{M}_h \rightarrow \mathcal{M}_r \) and 2) A humanoid whole-body control policy \( \pi_{\theta} \) that tracks the motion \( M_{\text{r}} \) and produces actions \( A_r \), i.e., \( \pi_{\theta}: \mathcal{M}_r \rightarrow \mathcal{A}_r \), while maintaining dynamic stability.

The complete pipeline is the composition: $\Pi(I) = \mathcal{A}_{\text{H-ACT}}(\mathcal{G}_{\text{H-GPT}}(I))$.
This formulation leverages human motion as an intermediate, semantically rich representation to bridge the gap between language and stable humanoid robot control.

% % % % % % % % % % % % % % % % % % % % %  Motion Planner % % % % % % % % % % % % % % % % % % % % % % % % % % % % % 
\begin{figure*}[t]
    \centering
    \includegraphics[width=1.0\textwidth]{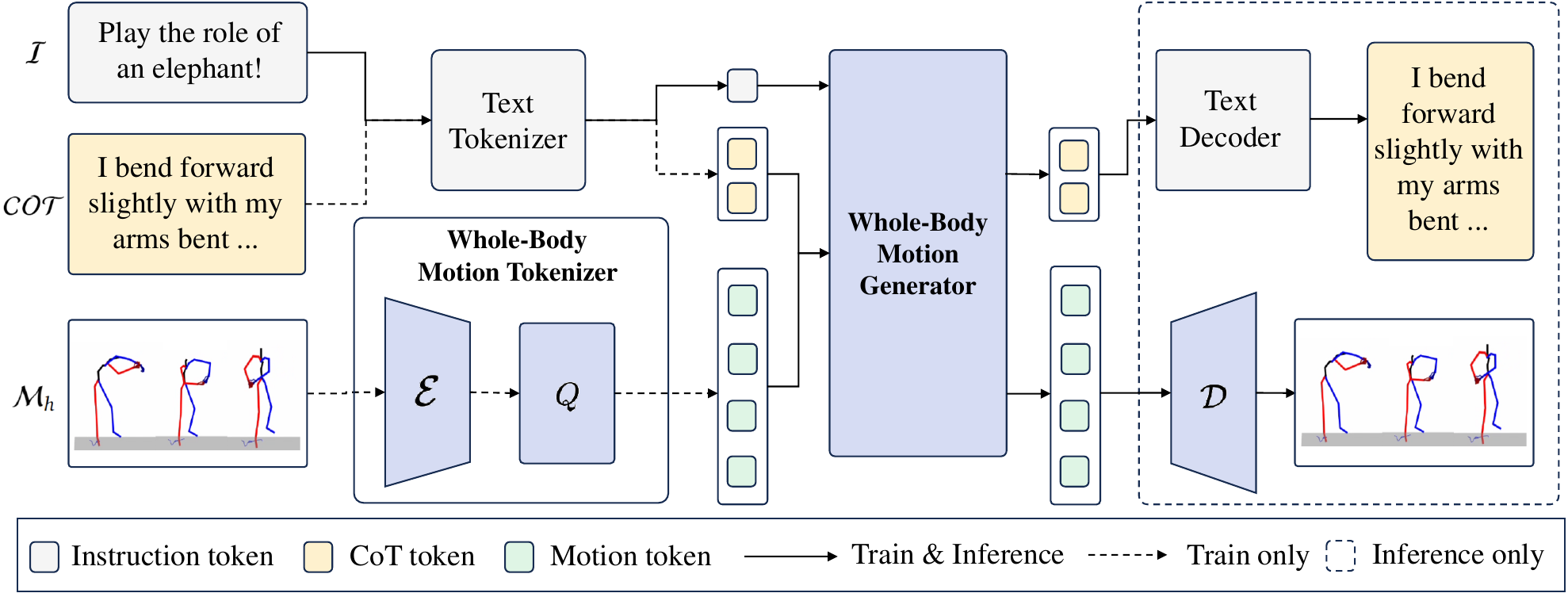}
    \caption{
Overview of the \texttt{H-GPT}. 
During \textbf{training} phase, data triplets <$\mathcal{I}$, $\mathcal{COT}$,$\mathcal{M}_h$> are tokenized and fed into the motion generator, where the motion sequences are encoded and discretized by the encoder $\mathcal{E}$ and the quantizer $Q$ of the whole-body motion tokenizer. 
During \textbf{inference} phase, the motion generator produces CoT and motion tokens given a specific instruction. 
These motion tokens are then decoded to a motion sequence by the decoder $\mathcal{D}$ of the tokenizer.
    }
    \label{fig:fromw1_hgpt}
\end{figure*}

\subsection{H-GPT}
\label{sec:hgpt}

The \texttt{H-GPT} module serves as the language-to-human-motion generator in the first stage of the \texttt{FRoM-W1} pipeline, learning the mapping \( \mathcal{G}_{\text{H-GPT}}: \mathcal{I} \rightarrow \mathcal{M}_{h} \). It synthesizes diverse and semantically consistent whole-body human motions from natural language instructions by employing a whole-body human motion tokenizer and an LLM-based auto-regressive motion generator. Figure~\ref{fig:fromw1_hgpt} illustrates the training and inference pipeline of \texttt{H-GPT}.

\subsubsection{Dataset Augmentation and Enrichment}
\label{sec:training_datasets}

Let $\mathcal{D}_{\text{raw}} = \{(I_i, M_{h,i})\}_{i=1}^{N}$ denote a raw dataset of $N$ samples, where $I_i \in \mathcal{I}$ is a natural language instruction and $M_{h,i} \in \mathcal{M}_h$ is a corresponding whole-body human motion sequence. While such paired data for humanoids is scarce, several large-scale datasets exist for whole-body human motion, e.g., HumanML3D~\cite{DBLP:conf/cvpr/GuoZZ0JL022} and Motion-X~\cite{DBLP:conf/nips/LinZLCZWZ23}. A key limitation is that the instruction space $\mathcal{I}$ in these datasets is often semantically narrow, consisting primarily of simple, concrete action descriptions (e.g., ``walk forward'', ``raise arms'') that lack the diversity and abstraction required for open-ended instructions like \textit{``Please play the role of an elephant!''}.

To bridge this semantic gap, we propose an LLM-augmented dataset transformation. We leverage a powerful large language model $\text{LLM}(\cdot)$ (specifically the OpenAI's GPT-4o) to enrich each sample. For each original pair $(I_i, M_{h,i})$, we generate a structured Chain-of-Thought (CoT)~\citep{DBLP:conf/nips/Wei0SBIXCLZ22} sequence $CoT_i \in \mathcal{COT}$ that decomposes and elaborates the instruction into explicit, temporally-aware motion primitives: $CoT_i = \text{LLM}(I_i, P)$,
where $P$ is a prompt template designed to elicit a step-by-step motion reasoning process. This yields an enriched dataset $\mathcal{D}_{\text{enriched}} = \{(I_i, CoT_i, M_i)\}_{i=1}^{N}$.

The CoT sequence $CoT_i$ serves as a semantic intermediary that aligns high-level or abstract language with low-level motion patterns. During training, the LLM-based auto-regressive generation model can learn the distribution $p(\mathcal{M}_h, \mathcal{COT} | \mathcal{I}) = p(\mathcal{M}_h | \mathcal{I}, \mathcal{COT}) \cdot p(\mathcal{COT} | \mathcal{I})$ by aligning the CoT sequence space $\mathcal{C}$ with corresponding human motion sequence space $\mathcal{M}_h$. During inference, for a novel instruction $I_{\text{new}}$, the model first generates a CoT $CoT_{\text{new}}$ via its internal reasoning capabilities, which then conditions the auto-regressive generation of the motion sequence $M_{\text{new}}$. 
This approach efficiently leverages the intrinsic language comprehension and generation capabilities of LLM base models to extend the original instruction space \( \mathcal{I} \) to a CoT space \( \mathcal{COT} \) that is more directly aligned with the target human motion space \( \mathcal{M}_h \), enabling the whole-body human motion generator robust generalization to complex, open-ended commands.

\subsubsection{Whole-Body Human Motion Tokenizer}
\label{sec:hgpt_tokenizer}

In order to achieve language-guided whole-body human motion generation based on the comprehension and sequence modeling capabilities of existing LLMs such as the LLaMA model~\citep{DBLP:journals/corr/abs-2407-21783}, we first train a whole-body human motion sequence tokenizer using VQ-VAE technology~\citep{DBLP:conf/nips/OordVK17}. This tokenizer represents a human motion sequence \(M_h = \{\mathbf{p}_{h,1}, \dots, \mathbf{p}_{h,T}\} \in \mathbb{R}^{T \times D}\) as a sequence of discrete motion tokens \(\hat{C} = \{\hat{c}_1, \dots, \hat{c}_L\} \in \hat{\mathcal{C}}\), where \(\mathbf{p}_{h,i} \in \mathbb{R}^D\) denotes the pose representation of the human at time \(i\), and \(\hat{c}_i \in \{1, \dots, K\}\) refers to the \(i\)-th token code id, aligning its format with that of linguistic tokens for subsequent LLM processing.

The tokenizer performs discrete representation through encoding, quantization, and decoding steps.
An encoder $\mathcal{E}$ first performs spatio-temporal downsampling: $\mathcal{E}: \mathbb{R}^{T \times D} \rightarrow \mathbb{R}^{L \times d}$ maps the sequence $M_h$ to latent vectors $Z = \{\mathbf{z}_1, \dots, \mathbf{z}_L\}$, where \(L = \lfloor T/l \rfloor\) and \(l\) is the temporal downsampling factor.
A vector quantizer \(\mathcal{Q}\) discretizes each latent vector \(\mathbf{z}_i\) by mapping it to the nearest entry in a learnable codebook \(C = \{\hat{\mathbf{z}}_1, \dots, \hat{\mathbf{z}}_K\} \subset \mathbb{R}^{d}\), producing a corresponding codebook index $\hat{c}_i$. 
The operation is defined as
$ \hat{c}_i = \mathcal{Q}(\mathbf{z}_i) = \underset{k \in \{1, \dots, K\}}{\arg\min} \|\mathbf{z}_i - \hat{\mathbf{z}}_k\|_2 $.
The resulting sequence of codebook indices \(\hat{C} = \{\hat{c}_1, \dots, \hat{c}_L\}\) constitutes the discrete motion tokens. 
%
% The decoder subsequently uses the corresponding codebook vectors \(\{\hat{\mathbf{z}}_{\hat{c}_1}, \dots, \hat{\mathbf{z}}_{\hat{c}_L}\}\) for reconstruction.
%
Finally, a decoder \(\mathcal{D}\) reconstructs the original motion sequence from the corresponding codebook vectors, producing \(\tilde{M}_h = \mathcal{D}(\hat{\mathbf{z}}_{\hat{c}_1}, \dots, \hat{\mathbf{z}}_{\hat{c}_L}) = \{\hat{\mathbf{p}}_{h,1}, \dots, \hat{\mathbf{p}}_{h,T} \}\in \mathbb{R}^{T \times D}\).

To train this motion tokenizer, we follow common practice~\citep{DBLP:conf/eccv/GuoZWC22,DBLP:journals/corr/abs-2301-06052,DBLP:conf/nips/JiangCLYYC23} by using the weighted sum of three loss functions as the training objective: reconstruction loss $\mathcal{L}_{recon}$, commitment loss $\mathcal{L}_{commit}$, and codebook loss $\mathcal{L}_{codebook}$. We also employ exponential moving average (EMA) updates for the codebook and a code reset mechanism to mitigate codebook collapse~\citep{DBLP:conf/nips/RazaviOV19}. 
% Detailed architectural specifications and hyperparameter values are provided in Appendix~\ref{app:hgpt_arch_training}.

\subsubsection{Whole-Body Human Motion Generator}
\label{sec:hgpt_generator}

The whole-body human motion generator, as the core component of \texttt{H-GPT}, is responsible for the auto-regressive generation of a coherent motion token sequence conditioned on a natural language instruction. It learns the distribution \( p(\hat{\mathcal{C}}, \mathcal{COT} \mid \mathcal{I}) \), where \(\hat{\mathcal{C}}\) is the discrete motion token sequence space derived from the motion tokenizer (Section \ref{sec:hgpt_tokenizer}). The generator is based on pretrained LLMs with an expanded vocabulary \( V = V_{\text{text}} \cup V_{\text{motion}} \). Here, \( V_{\text{text}} \) is the original text token vocabulary of the base LLM, and \( V_{\text{motion}} \) is a set of \(K\) new tokens corresponding to the motion codebook indices. This unified representation enables seamless interleaving of text (instructions and CoT) and motion tokens within a single sequence, transforming motion generation into a standard next-token prediction task.

During training, as illustrated in Figure~\ref{fig:fromw1_hgpt}, the model processes sequences in the $\mathcal{D}_{\text{enriched}}$ structured as: ``[Instruction] \(I\) [Chain-of-Thought] \(CoT\) [Motion Tokens] \(\hat{C}\) ''. The entire sequence is tokenized using the combined vocabulary \(V\). The model is trained with a standard language modeling objective to maximize the likelihood of the target sequence (CoT and motion tokens) given the instruction. Let \(\Theta\) denote the frozen parameters of the pre-trained backbone LLM like the LLaMA-3.1 model~\cite{DBLP:journals/corr/abs-2407-21783}, and \(\Delta\Theta\) denote the parameters of Low-Rank Adaptation (LoRA) modules~\cite{DBLP:conf/iclr/HuSWALWWC22} injected into the attention layers. The training loss is:
\begin{equation}
\label{formula:lm_loss}
\mathcal{L}_{lm} = -\log p(\hat{C}, CoT \mid I) = -\sum_{i=1}^{T} \log p(token_i \mid token_{<i}, I; \Theta + \Delta\Theta),
\end{equation}
where \( token_{<i} \) represents all preceding tokens in the sequence and \( token_i \) is the current token to be predicted, and $T$ is the token sequence length. This objective allows the model to learn the mapping from language to motion through the intermediate CoT representation, leveraging the world knowledge and reasoning capabilities of the base LLM while efficiently adapting it to the motion domain via parameter-efficient fine-tuning.

During inference for a novel instruction \(I_{\text{new}}\), the model performs auto-regressive generation in two phases, mirroring the learned sequence structure. First, it generates the Chain-of-Thought \(CoT_{\text{new}}\) conditioned on \(I_{\text{new}}\). Subsequently, conditioned on both \(I_{\text{new}}\) and the generated \(CoT_{\text{new}}\), it generates the sequence of motion tokens \(\hat{C}_{\text{new}}\). The generated motion tokens \(\hat{C}_{\text{new}}\) are then passed to the decoder \(D\) of the pre-trained motion tokenizer to reconstruct the corresponding continuous human motion sequence \(\tilde{M}_h\). This generated motion is semantically aligned with the instruction and serves as the input for the subsequent \texttt{H-ACT} module for robot execution. 

% Additional details of the training and inference settings are provided in Appendix~\ref{app:hgpt_arch_training}.

\subsection{H-ACT}
\label{sec:hact}

The \texttt{H-ACT} module is designed to enable humanoid robots to stably execute the whole-body motion sequence \(\tilde{M}_h\) generated by the \texttt{H-GPT} module in the physical world.
It mainly consists of three modules: a retargeter module $R$ that converts whole-body human motion representations \(\tilde{M}_h\) into motion sequences for a specific robot structure \(\tilde{M}_r\); a policy module $\pi_{\theta}$ trained via reinforcement learning in a physical simulation environment to stably track the retargeted humanoid motion sequences under factors such as gravity; and a lightweight, modular sim2real policy deployment framework that efficiently deploys the simulation-trained policy to the physical robot.

\begin{figure}[t]
    \centering
    \includegraphics[width=0.97\textwidth]{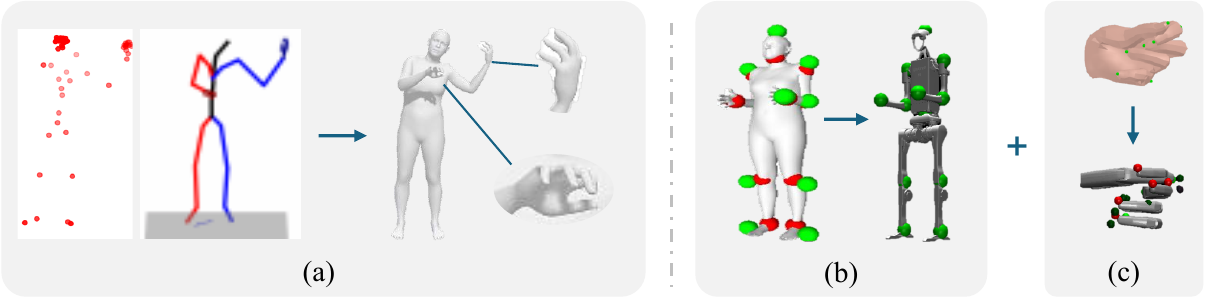}
    \caption{
        Overview of the human-to-humanoid motion retargeting pipeline.
        (a) We first convert the 3D coordinates of the full-body keypoints into the SMPL-X representation expressed in axis-angle format.
        (b) For body retargeting, we follow PHC to align global body poses, and additionally incorporate a rotation loss to compute wrist joint orientations. 
        (c) For hand retargeting, we formulate an optimization objective that aligns fingertip positions and solve it to obtain the final hand poses.
    }
    \label{fig:hact_retarget}
\end{figure}

\subsubsection{Human-to-Humanoid Whole-Body Motion Retargeter}
\label{sec:hact_retargeter}

To enable the physical humanoid robot to execute the whole-body human keypoint motion \(\tilde{M}_h\) generated by \texttt{H-GPT}, it is necessary to bridge the morphological gap between the human body and the robot structure. Our retargeting pipeline \(\mathcal{R}\) transforms \(\tilde{M}_h = \{\hat{p}_{h,1}, \dots, \hat{p}_{h,T}\}\) into a feasible robot joint-angle trajectory \(\tilde{M}_r = \{\hat{q}_{r,1}, \dots, \hat{q}_{r,T}\}\) through a two-stage process: first reconstructing a parametric human whole-body model SMPL-X~\citep{DBLP:conf/cvpr/PavlakosCGBOTB19} pose sequences \(\{\hat{q}_{h,1}, \dots, \hat{q}_{h,T}\}\) from the 3D human motion position sequences \(\tilde{M}_h\), and then optimizing the humanoid robot pose sequences $\tilde{M}_r$ to align with this reconstructed sequences.

We employ an inverse kinematics approach based on the HybrIK method~\citep{DBLP:conf/cvpr/LiXCBYL21} to reconstruct $t$-th SMPL-X pose $\hat{q}_{h,t}$ from the $t$-th 3D keypoint positions \(\hat{p}_{h,t} \in \mathbb{R}^{N \times 3}\) contained in \(\tilde{M}_h\). For each joint \(i\) with parent \(p(i)\), a rotation matrix \(R_i\) in the 3D Special Orthogonal Group $SO(3)$ is computed to align its template bone vector \(t_i\) with the observed direction \(p_i = \hat{p}_{h,t,i} - \hat{p}_{h,t,p(i)}\).
For joints with a single child, \(R_i\) is obtained via Rodrigues' formula~\citep{planning-kinematics}:
\begin{equation}
R_i = I + \sin\theta \, K + (1-\cos\theta) \, K^2,
\end{equation}
where \(K\) is the skew-symmetric matrix of the rotation axis \(k = \frac{t_i \times p_i}{\|t_i \times p_i\|}\) and \(\theta = \arccos(\frac{t_i \cdot p_i}{\|t_i\|\|p_i\|})\).
For joints with multiple children (e.g., the pelvis), the rotation is obtained by solving the orthogonal Procrustes problem:
\begin{equation}
\min_{R_i \in SO(3)} \left\| R_i \, T - P \right\|_F^2,
\end{equation}
where the template matrix \(T \in \mathbb{R}^{3 \times m}\) and the observed matrix \(P \in \mathbb{R}^{3 \times m}\) are constructed from the \(m\) child bone vectors. The solution, via singular value decomposition (SVD) \(P T^\top = U \Sigma V^\top\), is given by \(R_i = U V^\top\).
Rotation matrices are computed hierarchically from the root outward and subsequently converted into the SMPL-X pose parameter representation \(\hat{q}_{h,t}\), where each primitive represents the angle of a joint.

The reconstructed SMPL-X human whole-body motion is then mapped to the robot's joint space. For each frame \(t\), we solve for the robot joint angles \(\hat{q}_{r,t}\) that best align selected robot joints with their corresponding human joints. This is formulated as an optimization problem that minimizes positional and orientational discrepancies while respecting robot constraints:
\begin{equation}
\min_{\hat{q}_{r,t}} \sum_{(j_h, j_r) \in \mathcal{P}} \left[ \lambda_p \left\| \mathbf{p}\!\left( \text{FK}_r^{j_r}(\hat{q}_{r,t}) \right) - \hat{p}_{j_h,t} \right\|_2^2 \;+\; \lambda_r \, d\!\left( \mathbf{R}\!\left( \text{FK}_r^{j_r}(\hat{q}_{r,t}) \right),\, \hat{R}_{j_h,t} \right) \right] \;+\; \mathcal{R}_r(\hat{q}_{r,t}),
\end{equation}
where \(\text{FK}_r^{j_r}(\cdot)\) denotes the forward kinematics function for robot joint \(j_r\), \(\hat{p}_{j_h,t}\) and \(\mathbf{R}_{j_h,t}\) are the target position and orientation derived from the human SMPL-X pose $\hat{q}_{h,t}$, the human root transition $p_{root}$ and root orientation $o_{root}$, \(d(\cdot,\cdot)\) is the geodesic distance on \(SO(3)\), \(\mathcal{P}\) is the set of corresponding human-robot joint pairs, and \(\mathcal{R}_r(\hat{q}_{r,t})\) is a regularizer enforcing joint limits and other robot-specific feasibility constraints. 
The resulting sequence of joint angles \(\tilde{M}_r = \{\hat{q}_{r,1}, \dots, \hat{q}_{r,T}\}\), as well as the root information $p_{root,t}$ and $o_{root,t}$ serves as the reference motion for the downstream whole-body control policy.
In implementation, for body retargeting, we follow PHC~\cite{DBLP:conf/iccv/0002CWKX23} to align global body poses, and additionally incorporate a rotation loss to compute wrist joint orientations. For hand retargeting, we formulate an optimization objective that aligns fingertip positions and solve it to obtain the final hand poses.

\begin{figure}[tb]
    \centering
    \includegraphics[width=\textwidth]{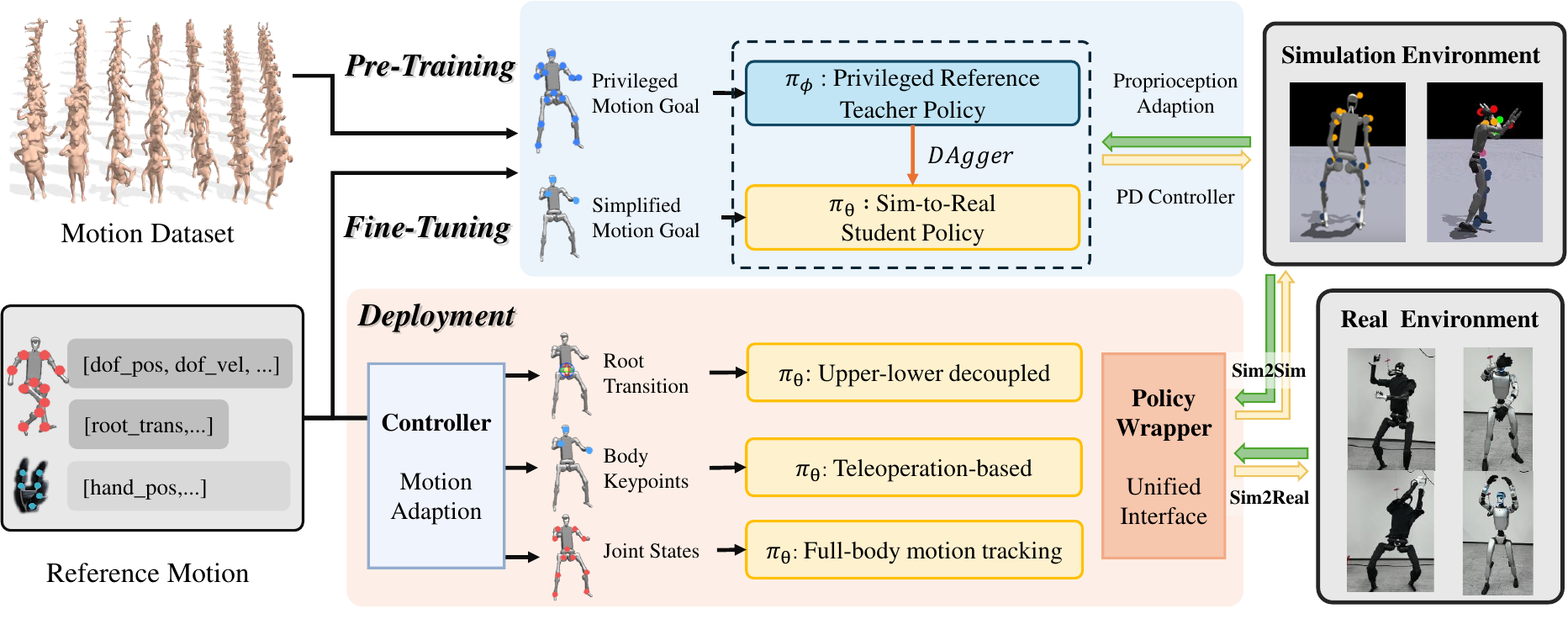}
    \caption{
        % \textcolor{red}{TODO; pli: add a draft version; edited;replace gr1t1; del SMPL}
        Overview of the \texttt{H-ACT}. 
        During the pre-training and fine-tuning phase, retargeted motion datasets are used in simulation to train a privileged teacher tracking policy. Through DAgger distillation, a sim-to-real student policy is trained with simplified motion goals;
        During the deployment phase, the generated humanoid motion is adapted to motion goals for different policy variants. A decoupled design with a unified interface enables seamless sim-to-sim transfer and sim-to-real deployment.
    }
    \label{fig:hact_imitator}
\end{figure}

\subsubsection{RL-based Motion Imitator: Pre-Training and Fine-Tuning}
\label{sec:hact_imitator}

As shown in Figure~\ref{fig:hact_imitator}, to enable the physical execution of a retargeted motion sequence \(\tilde{M}_r\) on a humanoid robot, we learn a robust whole-body control policy \(\pi_{\theta}\) using a two-stage RL framework in the IsaacGym simulator~\cite{DBLP:conf/nips/MakoviychukWGLS21}. 
The policy maps the robot's proprioceptive state \(s_t\) and a motion command \(c_t\)—comprising reference root position \(p_{\text{root},t}\), orientation \(o_{\text{root},t}\), and target joint angles \(\hat{q}_{r,t}\)—to motor actions \(a_t\).
While existing methods predominantly focus on pre-training a general motion controller~\citep{DBLP:conf/iros/HeLXZKLS24, he2024omnih2ouniversaldexteroushumantohumanoid, DBLP:journals/corr/abs-2505-02833, DBLP:journals/corr/abs-2508-08241}, we argue that fine-tuning the policy at inference time for a specific target motion can further improve tracking accuracy and robustness.

\paragraph{RL Pre-Training (RPT)} We first train a generalist policy using a large dataset of retargeted human motions \(\mathcal{D}_{\text{pt}} = \{ \tilde{M}_r^{(i)} \}_{i=1}^{N_{\text{pt}}}\) from AMASS~\citep{DBLP:conf/iccv/MahmoodGTPB19}. Following the OmniH2O approach~\citep{he2024omnih2ouniversaldexteroushumantohumanoid}, we adopt a teacher-student distillation paradigm.
A teacher policy \(\pi_{\phi}\), conditioned on privileged information (e.g., full state and future reference motions), is trained to maximize a cumulative reward across the entire pre-training dataset. The objective consists of three key components: an imitation reward \(r_t^{\text{im}}(\tilde{M}_r^{(i)})\) penalizing deviations from the reference motion in joint positions, velocities, and root trajectory; a stability reward \(r_t^{\text{stable}}\) encouraging balance and preventing falls; and optional regularization terms \(r_t^{\text{reg}}\). The objective is formalized as:
\begin{equation}
\max_{\phi} \; \mathbb{E}_{\tilde{M}_r^{(i)} \sim \mathcal{D}_{\text{pt}}, (s_t, a_t) \sim \pi_{\phi}} \left[ \sum_{t=0}^{T} \gamma^t \left( r_t^{\text{im}}(\tilde{M}_r^{(i)}) + \lambda_{\text{stable}} \cdot r_t^{\text{stable}} + \lambda_{\text{reg}} \cdot r_t^{\text{reg}} \right) \right],
\end{equation}
where \(\gamma \in (0,1]\) is a discount factor, and \(\lambda_{\text{stable}}\) and \(\lambda_{\text{reg}}\) are weighting coefficients for the stability and regularization rewards, respectively.

A deployable student policy \(\pi_{\theta}\) is then trained via supervised learning to mimic the teacher's actions, using only realistic onboard observations. The student's objective is to minimize the action discrepancy between its own output and the teacher's output, which serves as a guiding signal. This distillation process is formalized as:
\begin{equation}
\min_{\theta} \; \mathbb{E}_{\tilde{M}_r^{(i)} \sim \mathcal{D}_{\text{pt}}, c_t \sim \mathcal{U}(\tilde{M}_r^{(i)})} \left[ \bigl\| \pi_{\theta}(s_t, c_t) - \pi_{\phi}(s_t, c_t) \bigr\|^2 \right],
\end{equation}
where \(\mathcal{U}(\tilde{M}_r^{(i)})\) denotes unified goal state sampling from a motion trajectory. This objective arises from the core idea of policy distillation~\citep{DBLP:journals/jmlr/RossGB11}: the pre-trained, privileged teacher \(\pi_{\phi}\) provides high-quality target actions. The student \(\pi_{\theta}\) is trained to reproduce these target actions from its limited, real-world-compatible observation space \((s_t, c_t)\), thereby inheriting the teacher's performance while remaining practical for deployment.%
For example, during the training of the teacher policy, we track multiple key points of the human body on the reference human motion, while during the training of the student policy, we default to tracking only three key points: the head and both hands. To facilitate training and control, we adopt a direct retargeting approach for tracking human hand movements, without incorporating reinforcement learning into the training process.

\paragraph{RL Fine-Tuning (RFT)} During inference for a given target motion \(\tilde{M}_r^{\text{(tg)}}\), we fine-tune the pre-trained student policy \(\pi_{\theta}^{\text{(pt)}}\) exclusively on \(\tilde{M}_r^{\text{(tg)}}\), retaining the teacher-student structure. The teacher is fine-tuned with privileged access to \(\tilde{M}_r^{\text{(tg)}}\), and the student is subsequently distilled again via the same DAgger-based procedure, now specializing in the target motion. This process adapts the policy to the specific kinematic and dynamic characteristics of \(\tilde{M}_r^{\text{(tg)}}\), resulting in a policy \(\pi_{\theta}^{\text{(ft)}}\) that executes the motion with higher fidelity and stability than the generalist policy alone.

We use the PPO algorithm~\citep{DBLP:journals/corr/SchulmanWDRK17} to optimize these teacher policies with domain randomizations. Please note that as our default method, our pretrain-finetune paradigm employs a teacher-student training logic to articulate and validate the idea itself. But the entire pretrain-finetune paradigm is inherently compatible with the RL training of various common whole-body controllers~\citep{DBLP:journals/corr/abs-2505-02833, DBLP:journals/corr/abs-2508-08241}. For more details about the RPT and RFT, please refer to the Appendix~\ref{app:hact}.

\subsubsection{Modular Deployment Framework}
\label{sec:deployment_framework}

To enable robust real-world execution of trained policies, \texttt{H-ACT} further designs a modular framework \texttt{RoboJudo} that unifies a motion \textit{Controller} ($C$), an environment abstraction layer ($E$), and the policy module ($\Pi_{\theta}$) behind a consistent interface for motion commands $m_t$, robot state $s_t$, and actions $a_t$. 
This design not only supports the deployment of our default control methods across different robots but also seamlessly integrates various other forms of control paradigms through a plug-and-play approach.
The framework accommodates diverse whole-body policies by converting the canonical retargeted motion into policy-specific command formats:
\begin{itemize}
    \item \textbf{Upper-Lower Decoupled Policies} Split motion into upper-body joint targets and lower-body locomotion commands (root velocity, waist orientation)~\cite{DBLP:journals/corr/abs-2505-03738,DBLP:journals/corr/abs-2502-03206}.
    \item \textbf{Teleoperation-Based Policies} Condense motion into key joint or end-effector references for real-time tracking~\cite{DBLP:conf/iros/HeLXZKLS24,DBLP:journals/corr/abs-2505-02833}.
    \item \textbf{Whole-Body Tracking Policies} Use full joint and root trajectories as direct references for high-fidelity, per-motion tracking~\cite{DBLP:journals/corr/abs-2508-08241,DBLP:journals/corr/abs-2502-01143}.
\end{itemize}
This lightweight conversion enables a single generated motion \(\tilde{M}_r\) to drive multiple policy types. The Environment layer abstracts platform communication, while the Policy layer provides a unified execution interface $\pi_{\theta}: (s_t, m_t) \mapsto a_{t}$.

The modular design facilitates efficient sim-to-real transfer, runtime policy switching and composition, and streamlined integration of new robotic platforms. 
And this deployment framework provides a flexible and robust layer that translates generated humanoid motions \(\tilde{M}_r\) into stable, executable actions $\tilde{A}_r$ on physical humanoid robots.
The detailed system architecture is provided in Appendix~\ref{app:hact_robojudo}.

\section{Experiments}

% We conduct comprehensive experiments to evaluate \texttt{FRoM-W1} across three critical tasks:\hans{maybe change to answer 3 questions} (1) \textit{text-conditioned motion generation}, (2)\textit{simulation-based task execution} and (3) \textit{Full Pipeline deployment}. 
We conduct comprehensive experiments to answer three key questions about \texttt{FRoM-W1}:
\begin{itemize}
    \item \textbf{Q1}: Can the \texttt{H-GPT} model generate accurate and diverse whole-body human motions based on natural language instructions?
    \item \textbf{Q2}: Can the \texttt{H-ACT} module control humanoid robots to stably perform the motions generated by \texttt{H-GPT} in a simulation environment?
    \item \textbf{Q3}: How well does the entire framework perform when deployed on real-world humanoid robots?
\end{itemize}

%
% All experiments are conducted on a Unitree H1 humanoid robot with 35 DoFs, using NVIDIA A100 GPUs for training and an onboard Orin NX for real-time inference.

% For \textbf{Q2}, .For \textbf{Q3}, we tested the generalizability of \texttt{H-ACT} on Unitree H1\footnote{\url{https://www.unitree.com/cn/h1}} and Unitree G1\footnote{\url{https://www.unitree.com/cn/g1}} with an onboard Orin NX for real-time inference. Details of the setup are provided in ~\Cref{app:setup}.

\subsection{Instruction to Whole-Body Human Motion Generation}
To address \textbf{Q1}, we conducted a thorough and comprehensive evaluation of \texttt{H-GPT}. We first assessed \texttt{H-GPT}'s performance on text-to-motion benchmarks both quantitatively and qualitatively. Then, we tested the scalability of \texttt{H-GPT} by training it with more data. Additionally, we further examined the generalization ability of \texttt{H-GPT}'s CoT using abstract and complex instructions.

\paragraph{HumanML3D-X and $\delta$HumanML3D-X Benchmarks} Since there is no readily available benchmark for evaluating whole-body human motion generation capabilities that include hand movements, we extended the previous HumanML3D benchmark~\citep{DBLP:conf/cvpr/GuoZZ0JL022} to create two new benchmarks: HumanML3D-X and $\delta$HumanML3D-X. 

The HumanML3D Benchmark is originally constructed based on two mocap datasets, HumanAct12~\citep{DBLP:conf/mm/GuoZWZSDG020} and AMASS~\citep{DBLP:conf/iccv/MahmoodGTPB19}, and provide 3–4 textual descriptions for each motion clip. Models can be trained for text-to-motion generation using these text and motion pairs.
HumanML3D-X retains the text labels from HumanML3D and extends the original handless SMPL-format~\citep{DBLP:journals/tog/LoperM0PB15} motions to hand-inclusive SMPL-X-format~\citep{DBLP:conf/cvpr/PavlakosCGBOTB19} motions.
$\delta$HumanML3D-X further introduces two types of transformations based on HumanML3D-X: one involves altering the style of the original text labels (for example, from \textit{"a man kicks something or someone with his left leg."} to \textit{"So, this dude goes ahead and gives a good kick with his left leg to whatever's in his way."}), and the other adds noise perturbations to the original text labels (for example, from \textit{"a person is walking and then steps over something."} to \textit{"a person is wlaking then steps oveer something."}).
The HumanML3D-X benchmark is used to train and test models for generating whole-body human motion, by training on its training set and evaluating on its test set; while the $\delta$HumanML3D-X benchmark further evaluates models' generalization capabilities in response to language instructions, by testing models trained on HumanML3D-X directly on its test set.

We evaluated different methods using commonly employed scoring-model based metrics, including Fr{\'e}chet Inception Distance (FID), Retrieval Precision Top-$k$ (R Top-$k$, $k\in\{1,2,3\}$), Multimodal Distance (MM Dist), Diversity (DIV) and MultiModality (MModality)~\cite{DBLP:conf/cvpr/GuoZZ0JL022,DBLP:conf/cvpr/ChenJLHFCY23,DBLP:journals/corr/abs-2301-06052}.
(a) Motion quality: FID is our primary metric, used to measure the distance between the feature distributions of generated motions and ground-truth motions.
(b) Motion-Text matching: R Top-$k$ measures the accuracy of matching between text instructions and motions through retrieval methods. We use MM Dist to measure the distance between text and motions.
(c) Motion Diversity: DIV measures the diversity of motions by calculating the variance of their features. And MMoality measures the diversity of multiple motions generated under a fixed text description.

Following the workflow in \citet{DBLP:conf/cvpr/GuoZZ0JL022}, we retrained the corresponding text and motion feature extraction models as well as scoring models on these two benchmarks.
%

% More details of these benchmarks and metrics can be found in Appendix~\ref{app:benchmark_metric}. 

\begin{table*}[tb]
    \centering
    \begin{minipage}[t]{0.66\textwidth}
        \centering
        \vspace{0pt} % 确保顶部对齐
        \resizebox{\textwidth}{!}{
            \begin{tabular}{l|>{\columncolor[HTML]{F0F0FF}}ccccc}
                \toprule
                \textbf{Model} & \cellcolor[HTML]{F0F0FF}\textbf{FID $\downarrow$} & \textbf{R Top-1 $\uparrow$} & \textbf{MM Dist $\downarrow$} & \textbf{DIV $\rightarrow$} & \textbf{MModality $\uparrow$} \\
                \midrule
                \textbf{Real} & $\mathbf{0.000}^{\pm 0.00}$ & $\mathbf{0.393}^{\pm 0.01}$ & $\mathbf{3.862}^{\pm 0.01}$ & $\mathbf{9.811}^{\pm 0.10}$ & - \\
                \midrule
                T2M & \cellcolor[HTML]{F0F0FF}$2.078^{\pm 0.12}$ & $0.252^{\pm 0.01}$ & $4.967^{\pm 0.01}$ & $8.325^{\pm 0.13}$ & $\textbf{4.208}^{\pm 0.06}$ \\
                MotionDiffuse* & \cellcolor[HTML]{F0F0FF}$19.423^{\pm 0.08}$ & $0.100^{\pm 0.00}$ & $6.919^{\pm 0.01}$ & $5.183^{\pm 0.02}$ & $\underline{3.870}^{\pm 0.26}$ \\
                MLD & \cellcolor[HTML]{F0F0FF}$0.817^{\pm 0.05}$ & $\underline{0.357}^{\pm 0.01}$ & $\underline{4.189}^{\pm 0.04}$ & $\underline{9.549}^{\pm 0.10}$ & $3.224^{\pm 0.18}$ \\
                T2M-GPT & \cellcolor[HTML]{F0F0FF}$0.677^{\pm 0.06}$ & $\textbf{0.357}^{\pm 0.01}$ & $\textbf{4.182}^{\pm 0.02}$ & $9.260^{\pm 0.22}$ & $2.599^{\pm 0.13}$ \\
                \midrule
                $\text{H-GPT}_\text{w.o. CoT}$ & \cellcolor[HTML]{F0F0FF}$\textbf{0.229}^{\pm 0.03}$ & $0.333^{\pm 0.01}$ & $4.455^{\pm 0.03}$ & $\textbf{9.674}^{\pm 0.02}$ & $2.754^{\pm 0.34}$ \\
                $\text{H-GPT}$ & \cellcolor[HTML]{F0F0FF}$\underline{0.255}^{\pm 0.01}$ & $0.332^{\pm 0.00}$ & $4.513^{\pm 0.02}$ & $9.383^{\pm 0.18}$ & $3.256^{\pm 0.47}$ \\
                \bottomrule
            \end{tabular}
        }
        \vspace{-0.5em} % 微调表格与caption之间的距离
        \captionof{table}{\textbf{HumanML3D-X Benchmark}. Results are reported as mean ± 95\% confidence interval over three-time runs. The best results among generated motions are in \textbf{bold} and the second best results are \underline{underlined}. \colorbox[HTML]{F0F0FF}{FID} is our primary metric. *We found that MotionDiffuse is sensitive to parameters and difficult to achieve good results.}
        \label{tab:hgpt_main}
    \end{minipage}
    \hfill
    \begin{minipage}[t]{0.32\textwidth}
        \centering
        \vspace{0pt} % 确保顶部对齐
        \includegraphics[width=\textwidth, keepaspectratio=true]{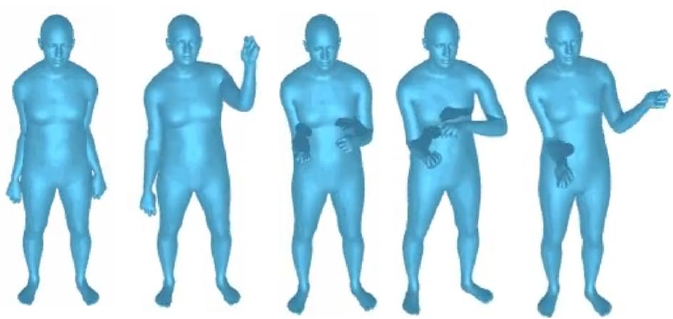} % height=0.9\textwidth
        \vspace{-2em} % 微调图片与caption之间的距离
        \captionof{figure}{Visualization of a motion sequence generated by the H-GPT model. The prompt is from the test set of the HumanML3D-X benchmark: \textit{``a person flings an object with their right hand then showcases something with both hands.''}}
        \label{fig:hgpt_vis}
    \end{minipage}
\end{table*}

\begin{table*}[tb]
    \centering
    \resizebox{1.0\textwidth}{!}{
    \begin{tabular}{l|>{\columncolor[HTML]{F0F0FF}}cccc|cccc}
        \toprule
        \multirow{2}{*}{\textbf{Model}} & \multicolumn{4}{c|}{\textbf{Instruction Rephrasing}} & \multicolumn{4}{c}{\textbf{Instruction Noising}} \\
        \cmidrule(lr){2-5} \cmidrule(lr){6-9}
        & \cellcolor[HTML]{F0F0FF}\textbf{FID $\downarrow$} & \textbf{R Top-1 $\uparrow$} & \textbf{DIV $\rightarrow$} & \textbf{MModality $\uparrow$} & \cellcolor[HTML]{F0F0FF}\textbf{FID $\downarrow$} & \textbf{R Top-1 $\uparrow$} & \textbf{DIV $\rightarrow$} & \textbf{MModality $\uparrow$} \\
        \midrule
        \textbf{Real} & \cellcolor[HTML]{F0F0FF}$\mathbf{0.000}^{\pm 0.000}$ & $\mathbf{0.367}^{\pm 0.001}$ & $\mathbf{8.863}^{\pm 0.054}$ & - & \cellcolor[HTML]{F0F0FF}$\mathbf{0.000}^{\pm 0.000}$ & $\mathbf{0.322}^{\pm 0.003}$ & $\mathbf{8.422}^{\pm 0.082}$ & - \\
        \midrule
        MLD & \cellcolor[HTML]{F0F0FF}$1.944^{\pm 0.168}$ & $0.165^{\pm 0.006}$ & $\textbf{8.357}^{\pm 0.113}$ & $4.075^{\pm 0.107}$ & \cellcolor[HTML]{F0F0FF}$2.256^{\pm 0.092}$ & $0.138^{\pm 0.008}$ & $\textbf{8.509}^{\pm 0.111}$ & $4.086^{\pm 0.096}$ \\
        T2M-GPT & \cellcolor[HTML]{F0F0FF}$2.043^{\pm 0.025}$ & $\textbf{0.234}^{\pm 0.001}$ & $7.578^{\pm 0.194}$ & $2.698^{\pm 0.169}$ & \cellcolor[HTML]{F0F0FF}$1.610^{\pm 0.001}$ & $\mathbf{0.183}^{\pm 0.005}$ & $6.886^{\pm 0.163}$ & $2.897^{\pm 0.353}$ \\
        \midrule
        $\text{H-GPT}_\text{w.o. CoT}$ & \cellcolor[HTML]{F0F0FF}$0.455^{\pm 0.021}$ & $0.210^{\pm 0.004}$ & $8.201^{\pm 0.201}$ & $4.240^{\pm 0.184}$ & \cellcolor[HTML]{F0F0FF}$0.830^{\pm 0.050}$ & $0.177^{\pm 0.008}$ & $7.677^{\pm 0.116}$ & $4.506^{\pm 0.389}$ \\
        $\text{H-GPT}$ & \cellcolor[HTML]{F0F0FF}$\textbf{0.355}^{\pm 0.007}$ & $0.225^{\pm 0.008}$ & $8.240^{\pm 0.207}$ & $\textbf{4.533}^{\pm 0.349}$ & \cellcolor[HTML]{F0F0FF}$\textbf{0.602}^{\pm 0.055}$ & $\textbf{0.192}^{\pm 0.006}$ & $7.788^{\pm 0.184}$ & $\textbf{4.642}^{\pm 0.367}$ \\
        \bottomrule
    \end{tabular}
    }
    \caption{\textbf{$\delta$HumanML3D-X Benchmark}. Results are reported as mean ± 95\% confidence interval over three-time runs. The best results among generated motions are in \textbf{bold}. \colorbox[HTML]{F0F0FF}{FID} is our primary metric. These models were trained on HumanML3D-X and evaluated on this benchmark.}
    \label{tab:hgpt_delta}
\end{table*}

\paragraph{Baselines}
Our baseline methods primarily include four mainstream approaches, falling into two categories: autoregressive models~\citep{DBLP:conf/nips/VaswaniSPUJGKP17} and diffusion models~\citep{DBLP:conf/nips/HoJA20}.
\begin{itemize}
    \item \textbf{T2M}~\citep{DBLP:conf/cvpr/GuoZZ0JL022} employs an autoencoder to encode motion features, and uses a recurrent neural network to generate motions of the corresponding length based on the estimated motion duration and textual description features.
    \item \textbf{MotionDiffuse}~\citep{DBLP:journals/pami/ZhangCPHGYL24} utilizes a linear-attention-based diffusion model for motion generation with a Transformer-based text encoder to extract the text embedding as the condition.
    \item \textbf{MLD}~\citep{DBLP:conf/cvpr/ChenJLHFCY23} applies the CLIP model~\citep{DBLP:conf/icml/RadfordKHRGASAM21} as an extractor for text features and utilizes a latent diffusion model~\citep{DBLP:conf/cvpr/RombachBLEO22} to generate motions.
    \item \textbf{T2M-GPT}~\citep{DBLP:journals/corr/abs-2301-06052} models the motion space with a stand-alone Transformer~\cite{DBLP:conf/nips/VaswaniSPUJGKP17} decoder and the text embeddings are also obtained from an off-the-shelf CLIP model.
\end{itemize}
These works were previously trained based on the HumanML3D dataset. We meticulously retrained these models on the HumanML3D-X training data using the open-source code from these works. 
% For further implementation details on the baselines, please refer to the Appendix~\ref{app:baselines}.

\paragraph{Implementation Details} In this experiment, we represented hand-inclusive whole-body motion sequences using position, velocity and rotation information, and represent rotations using 6D continuous representation~\citep{DBLP:conf/cvpr/ZhouBLYL19}. We primarily implemented two designs of \texttt{H-GPT}: one is \texttt{H-GPT} without CoT ($\text{H-GPT}_\text{w.o. CoT}$), and the other is \texttt{H-GPT} with CoT ($\text{H-GPT}$).
Both of them utilized the same motion tokenizer we trained, $\mathrm{K} \in \mathcal{R}^{512\times512}$, with the temporal downsampling factor $l$ set to 4.
Both \texttt{H-GPT} models are based on the 8B LLAMA-3.1 base model~\citep{DBLP:journals/corr/abs-2407-21783} and were further pretrained using LoRA~\citep{DBLP:conf/iclr/HuSWALWWC22} with a rank of 32, a scaling factor of 16, and a dropout rate of 0.05 applied to all linear projection layers, while the embedding and language modeling head layers remain fully trainable.
We used an 8-GPU NVIDIA machine for model training. For training the VQ-VAE, we set the batch size to 512 and trained for 30,000 epochs. For training the \texttt{H-GPT} models, we set the mini-batch size to 8 and trained for 600 epochs.
For more details on constructing the CoT data, please refer to Appendix~\ref{app:hgpt_dataset}. 
% For further details on the model architecture and training, please refer to Appendix~\ref{app:hgpt_arch_training}.

% \textcolor{red}{TBC} We first train the whole-body motion tokenizer and make abundant ablations to select the best setting. Details of the metrics and results are provided in appendix.  
% %
% Then, based on the best motion tokenizer, we train the \texttt{H-GPT} and evaluate its performance on virtual motion generation with hand feature(i.e., 623-dimension feature per frame) on 1024 randomly chosen test samples from~\cite{DBLP:conf/nips/LinZLCZWZ23}. We also compare our method with various baselines. 

% Moreover, to comprehensively assess the generative model, we evaluate it with both unretargeted motion features (623-dimensional feature per frame; results shown in~\Cref{tab:unretarget_res}) and motions retargeted to the Unitree H1 robot (40-dimensional feature per frame).
% Respectively, we train scoring models to compute the metric for both scenarios following~\cite{DBLP:conf/cvpr/GuoZZ0JL022}\footnote{\url{https://github.com/EricGuo5513/text-to-motion}}. 

\paragraph{Results} The experimental results on the HumanML3D-X benchmark are shown in Table~\ref{tab:hgpt_main}. We observe that our $\text{H-GPT}_\text{w.o. CoT}$ and \texttt{H-GPT} models achieve significant improvement on the main metric FID, reducing it from 0.677 of T2M-GPT to 0.229 and 0.255. This indicates that our \texttt{H-GPT} models are capable of generating human whole-body motions that better align with the textual descriptions. 
Meanwhile, our $\text{H-GPT}_\text{w.o. CoT}$ model also achieves the best result 9.674 on the motion diversity metric DIV.
Regarding the two metrics that measure the correlation between motion and text, R Top-1 and MM Dist, we observed in our experiments that T2M-GPT and our H-GPT models often result in their degradation when FID improves.
We believe this may be because the training loss of the model is primarily designed to fit the motion distribution in the training dataset, without imposing explicit constraints on the representation correlation between motion and text. This leads the model to more easily generate common and less distinguishable motions during the fitting process.
Regarding the MModality metric, we found that even though the fitting performance of models like T2M and MotionDiffuse is relatively poor, the diversity of motions generated from the same text tends to be higher for this metric.
We visualize an example of a human motion generated by the H-GPT model in Figure~\ref{fig:hgpt_vis}. 

\begin{table*}[tbp]
    \centering
    \resizebox{1.0\textwidth}{!}{
    \begin{tabular}{l|>{\columncolor[HTML]{F0F0FF}}ccccccc}  % 在FID列前添加背景色
        \toprule
        \multirow{2}{*}{\textbf{Model}} & \multirow{2}{*}{\cellcolor[HTML]{F0F0FF}\textbf{FID $\downarrow$}} & \multicolumn{3}{c}{\textbf{R Precision}} & \multirow{2}{*}{\textbf{MM Dist $\downarrow$}} & \multirow{2}{*}{\textbf{DIV $\rightarrow$}} & \multirow{2}{*}{\textbf{MModality $\uparrow$}} \\
        \cmidrule(lr){3-5}
        & & \textbf{Top-1 $\uparrow$} & \textbf{Top-2 $\uparrow$} & \textbf{Top-3 $\uparrow$} & & & \\
        \midrule
        \textbf{Real} & $\mathbf{0.000}^{\pm 0.000}$ & $\mathbf{0.393}^{\pm 0.005}$ & $\mathbf{0.573}^{\pm 0.005}$ & $\mathbf{0.677}^{\pm 0.004}$ & $\mathbf{3.862}^{\pm 0.009}$ & $\mathbf{9.811}^{\pm 0.096}$ & - \\
        \midrule
        $\text{H-GPT}_\text{w.o. CoT}$ & \cellcolor[HTML]{F0F0FF}$\textbf{0.229}^{\pm 0.029}$ & $\textbf{0.333}^{\pm 0.007}$ & $\textbf{0.490}^{\pm 0.003}$ & $\textbf{0.588}^{\pm 0.003}$ & $\textbf{4.455}^{\pm 0.033}$ & $\textbf{9.674}^{\pm 0.021}$ & $2.754^{\pm 0.335}$ \\
        $\text{H-GPT}$ & \cellcolor[HTML]{F0F0FF}$\underline{0.255}^{\pm 0.008}$ & $\underline{0.332}^{\pm 0.004}$ & $0.481^{\pm 0.004}$ & $0.573^{\pm 0.001}$ & $4.513^{\pm 0.019}$ & $9.382^{\pm 0.182}$ & $\textbf{3.256}^{\pm 0.465}$ \\
        \midrule
        $\text{H-GPT++}_\text{w.o. CoT}$ & \cellcolor[HTML]{F0F0FF}$0.337^{\pm 0.011}$ & $0.323^{\pm 0.005}$ & $0.482^{\pm 0.005}$ & $0.581^{\pm 0.004}$ & $\underline{4.494}^{\pm 0.054}$ & $9.240^{\pm 0.104}$ & $2.664^{\pm 0.115}$ \\
        $\text{H-GPT++}$ & \cellcolor[HTML]{F0F0FF}$0.312^{\pm 0.054}$ & $0.327^{\pm 0.004}$ & $\underline{0.486}^{\pm 0.010}$ & $\underline{0.583}^{\pm 0.012}$ & $4.494^{\pm 0.066}$ & $\underline{9.411}^{\pm 0.062}$ & $\underline{3.153}^{\pm 0.241}$ \\
        \bottomrule
    \end{tabular}
    }
    \caption{\textbf{HumanML3D-X Benchmark Results with Motion-X Training Data}. Performance comparison of different motion generation models against real motion data. Results are reported as mean ± 95\% confidence interval over multiple runs. The best results among generated motions are in \textbf{bold} and the second best results are \underline{underlined}. Real motion data serves as the ground truth reference.}
    \label{tab:hgpt_motionx}
\end{table*}

Moreover, we found on the HumanML3D-X benchmark that the $\text{H-GPT}_\text{w.o. CoT}$ model performs relatively better than the $\text{H-GPT}$ model (FID 0.229 vs. 0.255). We think this is partly because the training and test sets of HumanML3D-X have relatively consistent distributions, making it easier to learn the mapping between text and motion. In contrast, the introduction of CoT, due to the addition of a certain length of text, makes the fitting slightly more challenging.
Furthermore, based on our newly extended $\delta$HumanML3D-X benchmark on Table~\ref{tab:hgpt_delta}, it can be observed that compared to the baselines, both H-GPT models demonstrate better generalization capabilities in language understanding. Additionally, the introduction of CoT further enhances their ability to comprehend instructions and generate motions under the scenarios of instruction rephrasing (FID 0.455 $\rightarrow$ 0.355) and instruction noising (FID 0.830 $\rightarrow$ 0.602).
% For more results and analysis, please refer to the Appendix~\ref{app:hgpt_result}.

% It can be seen that our H-GPT model exhibits better generation effects compared to the baseline method.

\begin{figure}[tb]
    \centering
    % 第一组图片
    \begin{subfigure}{0.24\textwidth}
        \includegraphics[width=\linewidth]{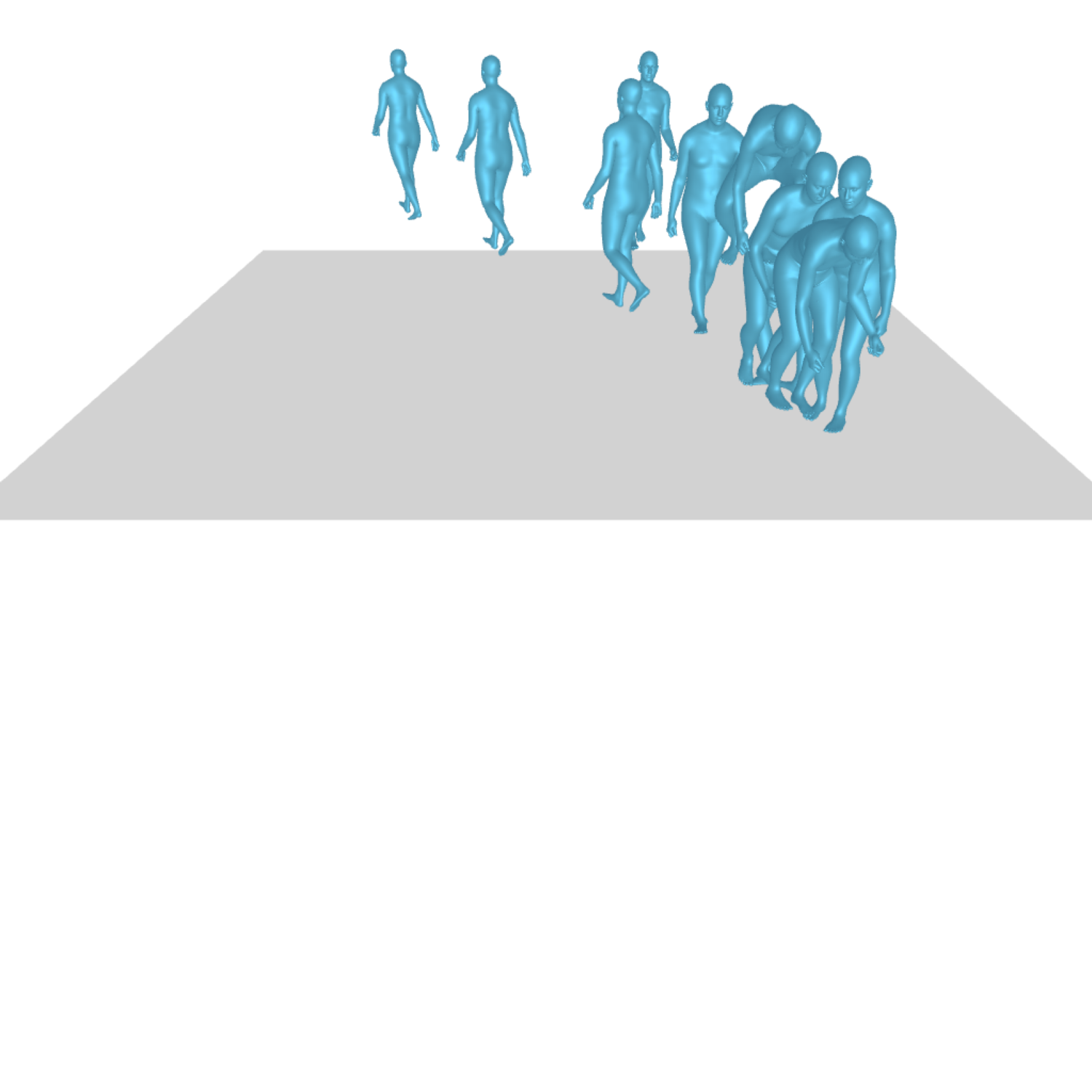}
        \caption{$\text{H-GPT}_\text{w.o. CoT}$}
        \label{fig:hgpt_cot_1a}
    \end{subfigure}\hfill
    \begin{subfigure}{0.24\textwidth}
        \includegraphics[width=\linewidth]{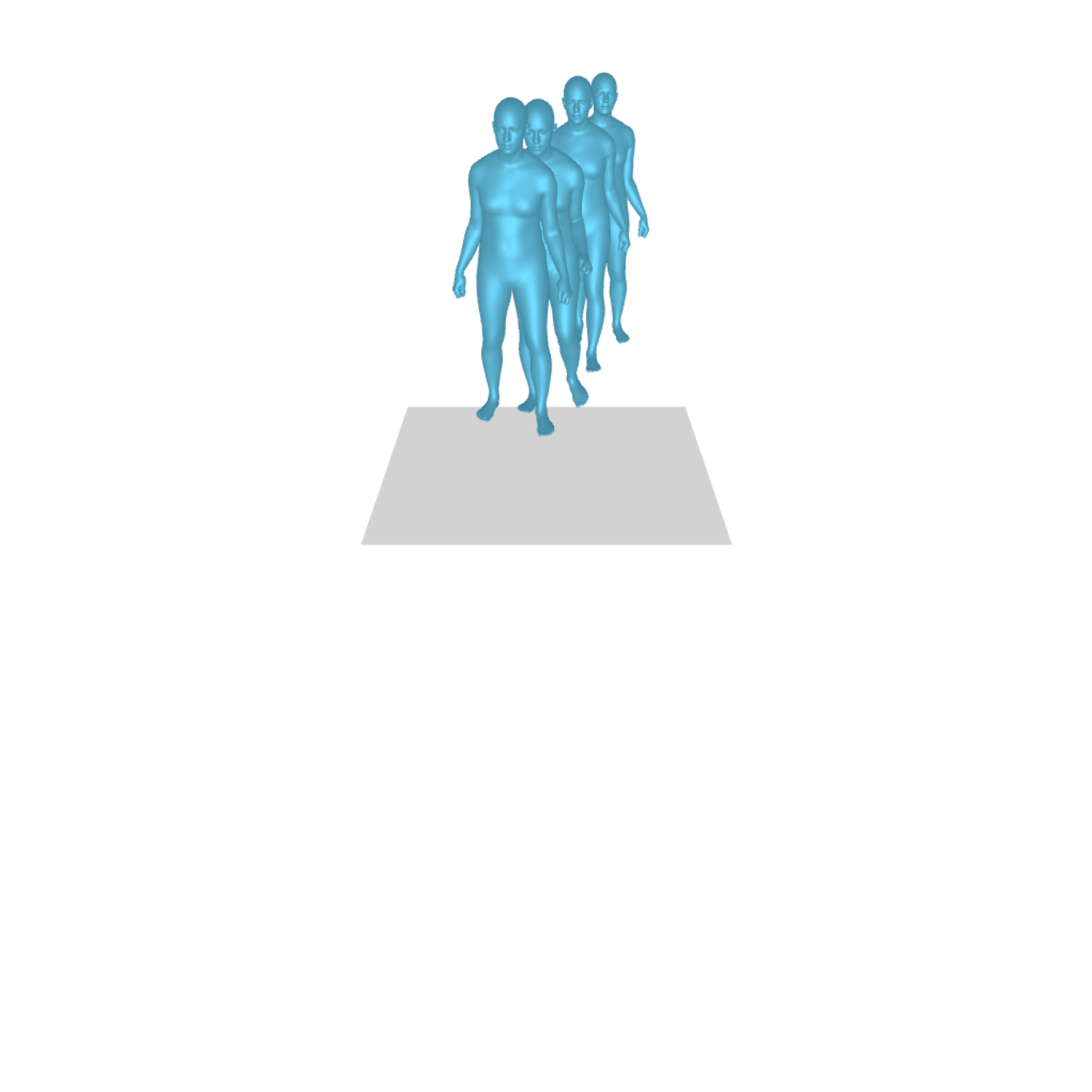}
        \caption{$\text{H-GPT}$}
        \label{fig:hgpt_cot_1b}
    \end{subfigure}\hfill
    \begin{subfigure}{0.24\textwidth}
        \includegraphics[width=\linewidth]{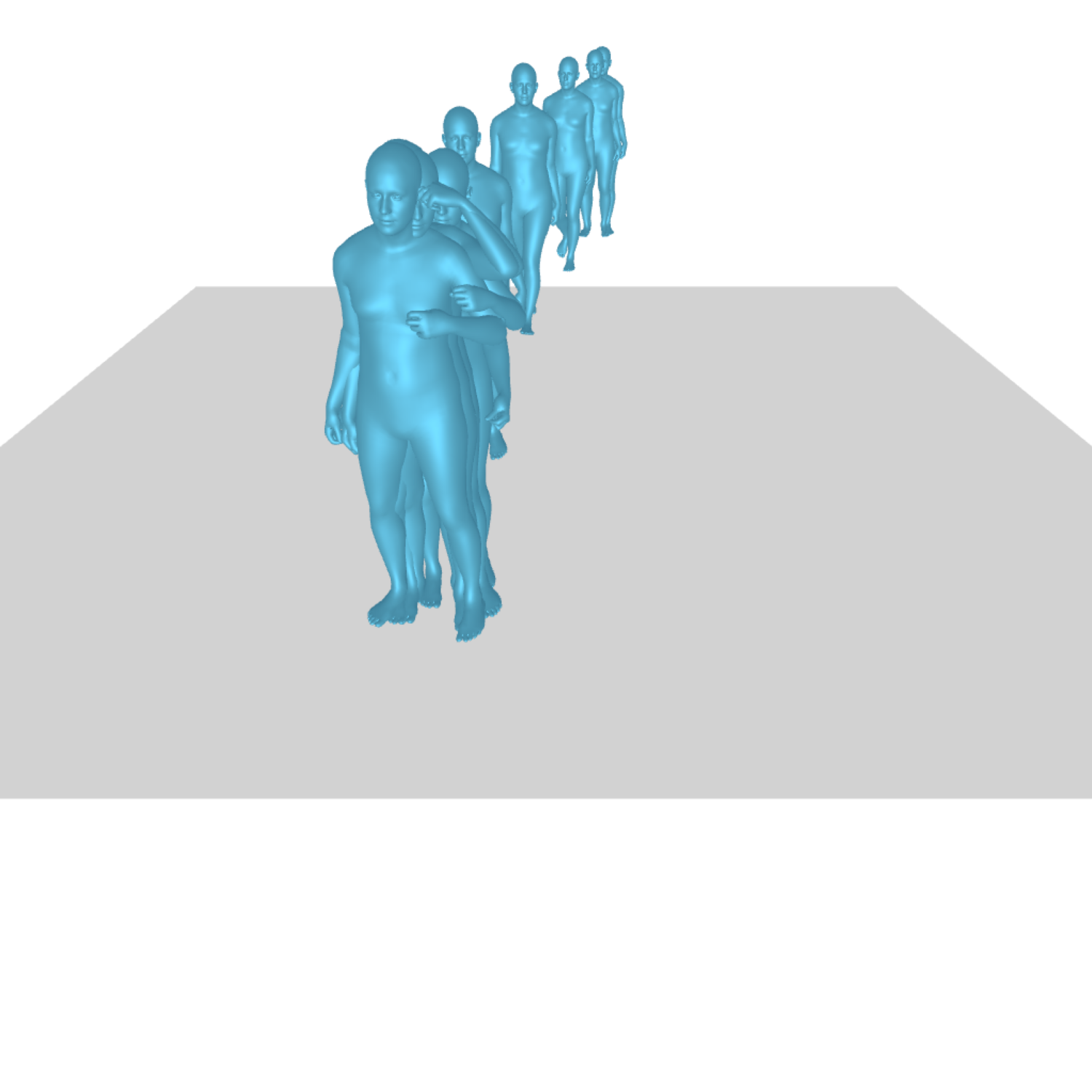}
        \caption{$\text{H-GPT++}_\text{w.o. CoT}$}
        \label{fig:hgpt_cot_1c}
    \end{subfigure}\hfill
    \begin{subfigure}{0.24\textwidth}
        \includegraphics[width=\linewidth]{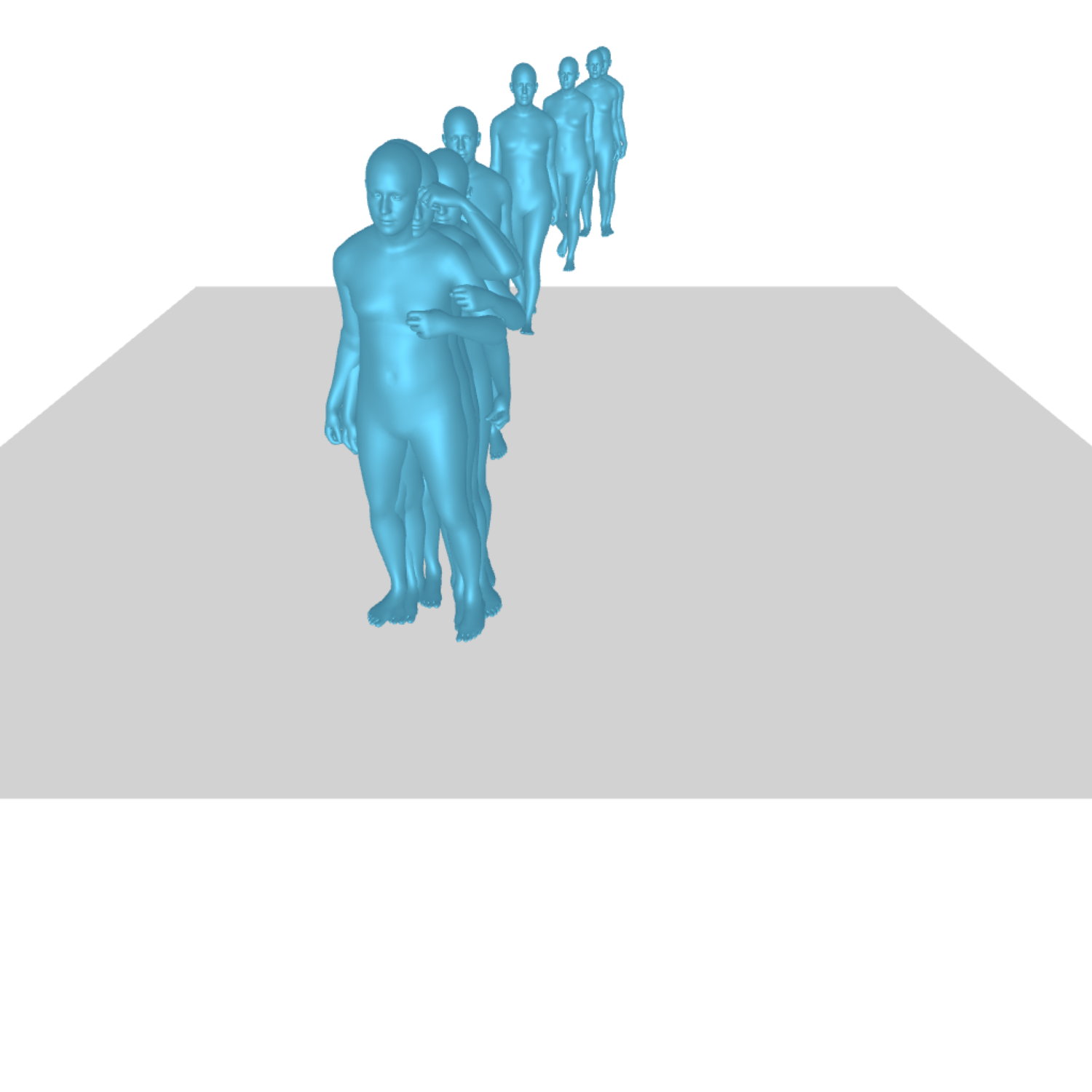}
        \caption{$\text{H-GPT++}$}
        \label{fig:hgpt_cot_1d}
    \end{subfigure}
    
    \vspace{0.2cm} % 行间距，可调整
    
    % 第二组图片
    \begin{subfigure}{0.24\textwidth}
        \includegraphics[width=\linewidth]{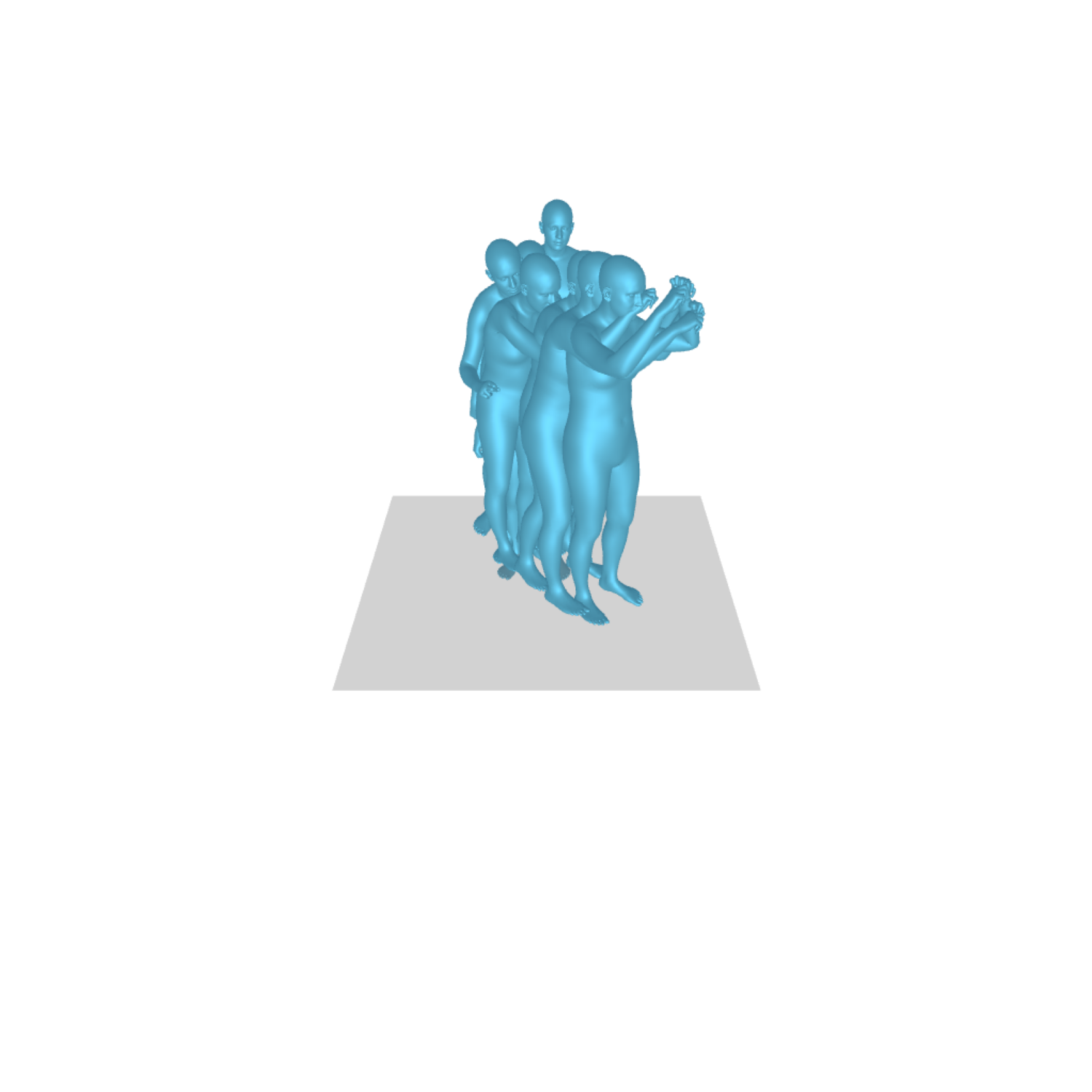}
        \caption{$\text{H-GPT}_\text{w.o. CoT}$}
        \label{fig:hgpt_cot_2a}
    \end{subfigure}\hfill
    \begin{subfigure}{0.24\textwidth}
        \includegraphics[width=\linewidth]{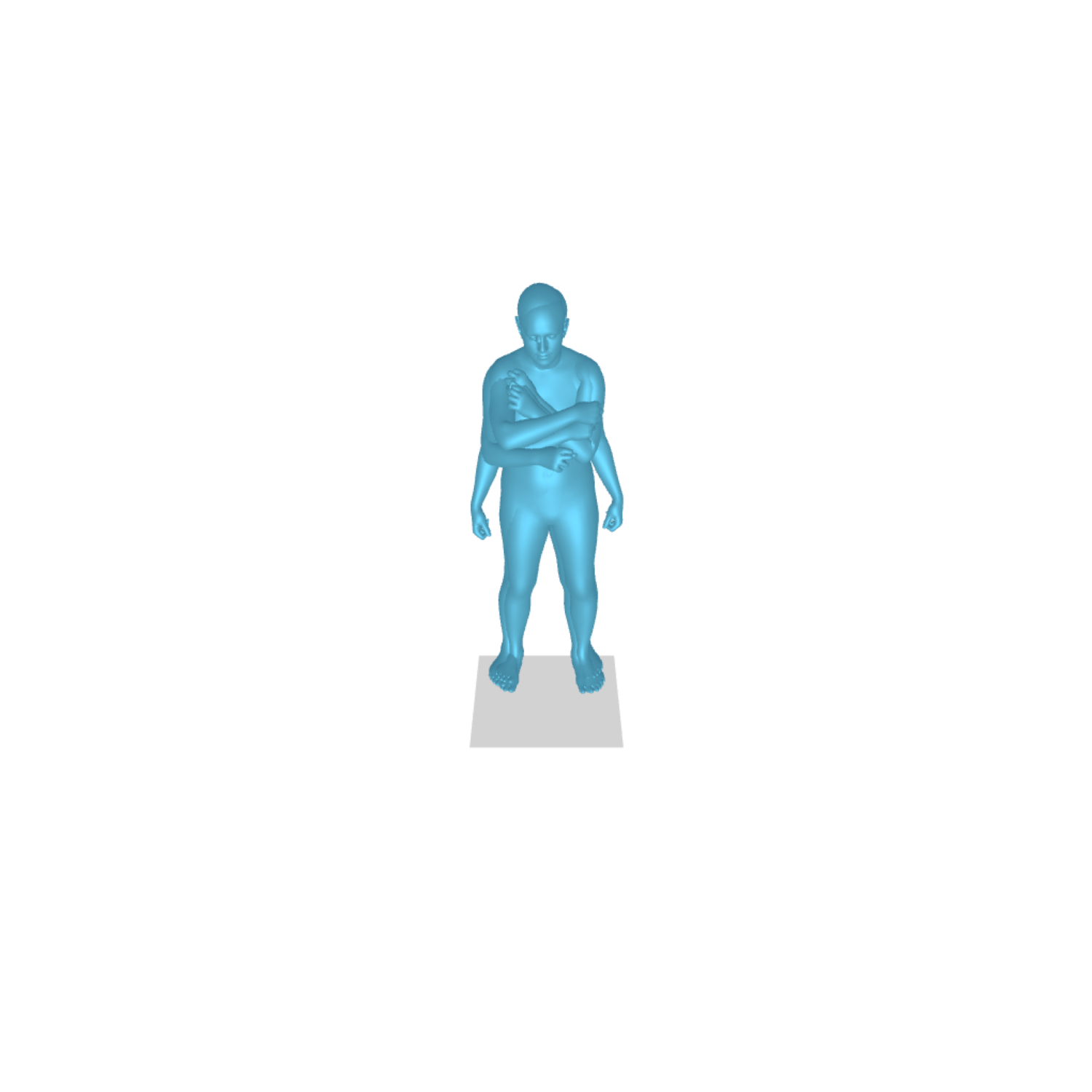}
        \caption{$\text{H-GPT}$}
        \label{fig:hgpt_cot_2b}
    \end{subfigure}\hfill
    \begin{subfigure}{0.24\textwidth}
        \includegraphics[width=\linewidth]{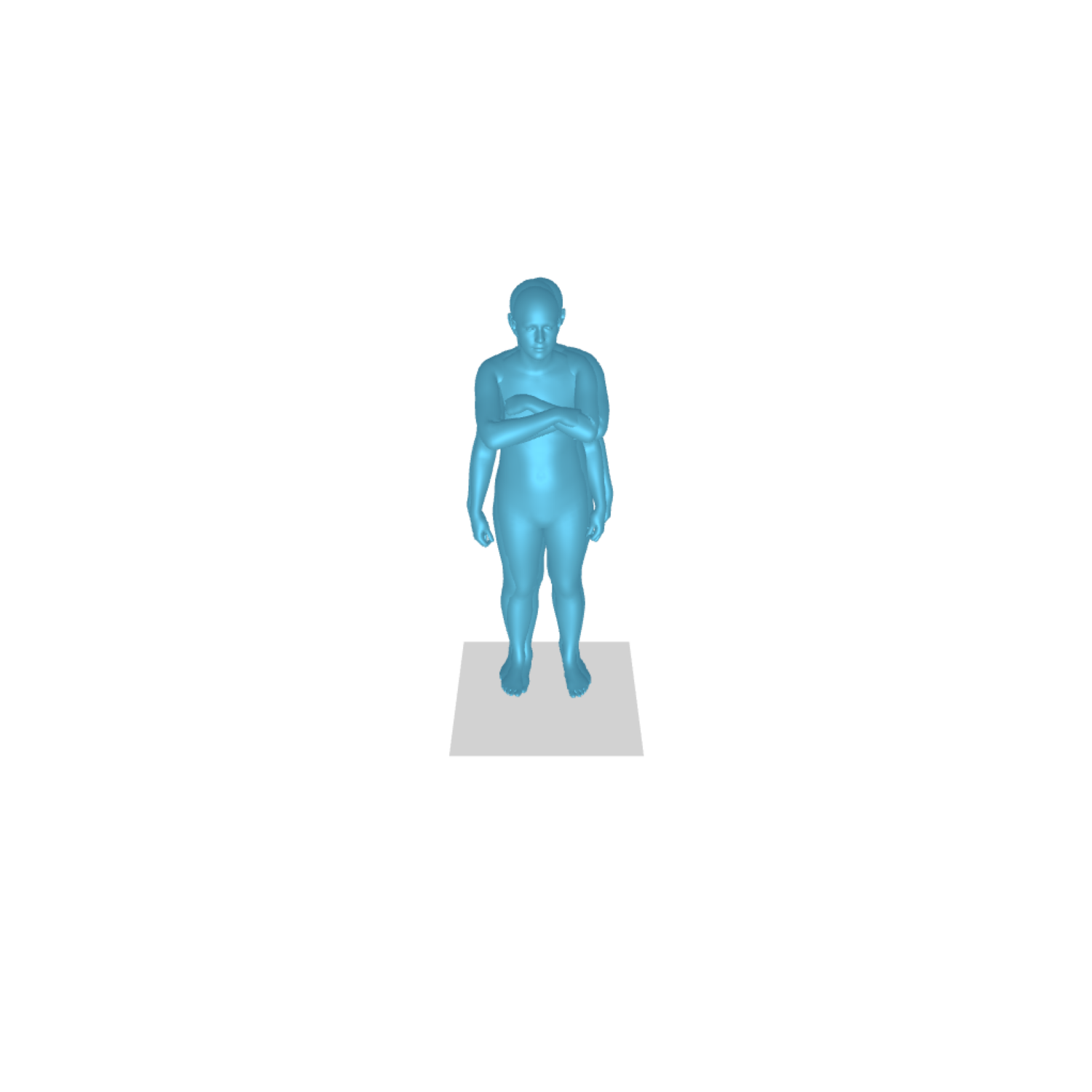}
        \caption{$\text{H-GPT++}_\text{w.o. CoT}$}
        \label{fig:hgpt_cot_2c}
    \end{subfigure}\hfill
    \begin{subfigure}{0.24\textwidth}
        \includegraphics[width=\linewidth]{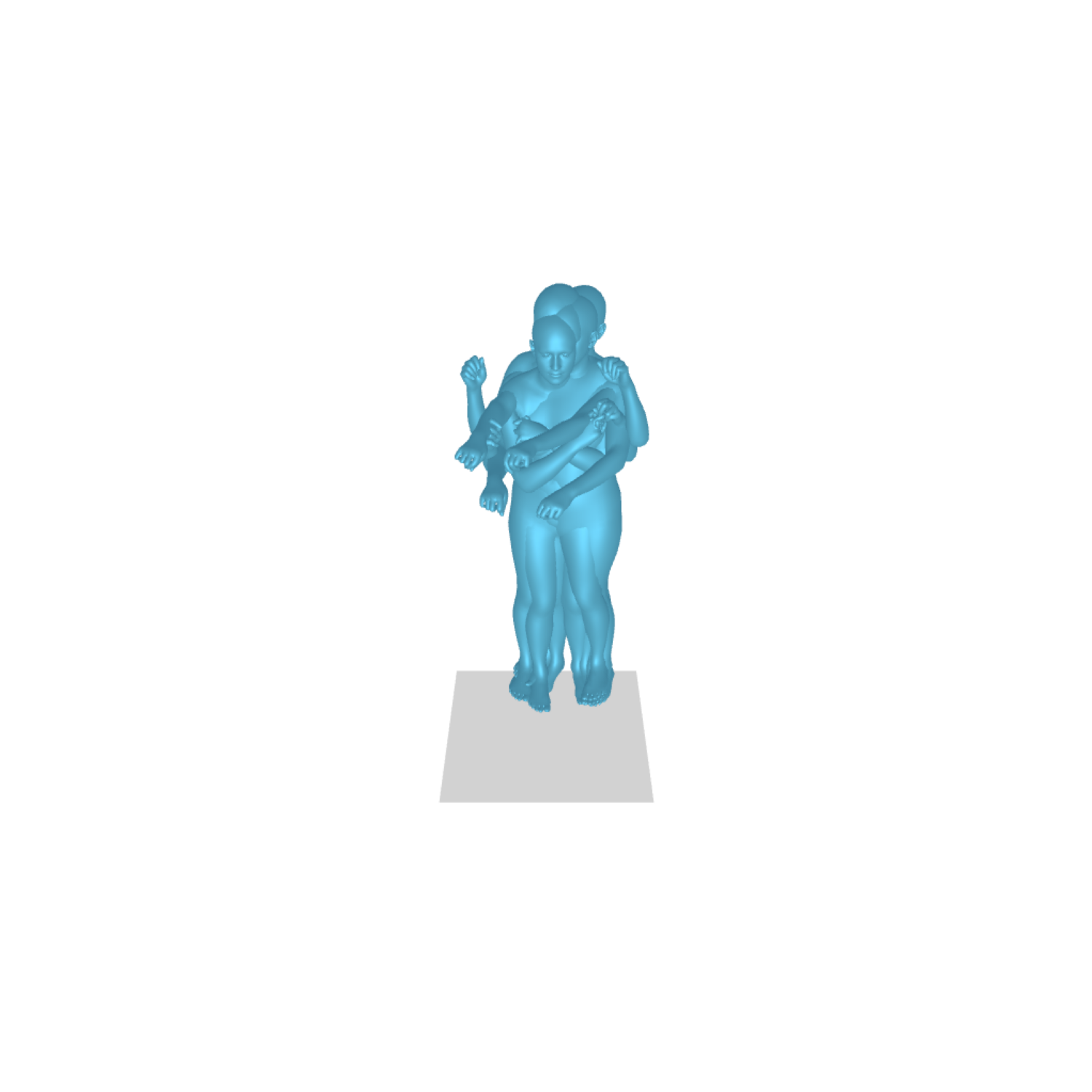}
        \caption{$\text{H-GPT++}$}
        \label{fig:hgpt_cot_2d}
    \end{subfigure}
    
    \caption{Complex (first row) and abstract (second row) cases visualization of different \texttt{H-GPT} models. Complex Prompt: \textit{``A person walks forward, stops, and stands drinking something.''}. Abstract Prompt: \textit{``Shiver and hug yourself tightly to communicate feeling cold or scared.''}}
    \label{fig:hgpt_cot}
\end{figure}

\paragraph{Data Scaling with Motion-X} As mentioned before, we argue that massive human data, especially human motion data extracted from vast video collections, has the potential to enhance the generalization of whole-body motion generation tasks for humanoid robots.
Therefore, we further conducted data processing and model training on Motion-X~\citep{DBLP:conf/nips/LinZLCZWZ23}, the largest publicly available human motion dataset that includes hand movements, including 81084 annotated human motion clips.
Motion-X encompasses data such as HumanML3D-X from motion capture and has additionally estimated a large volume of whole-body human motion poses from extensive video sources, with text labels generated using large vision-language models (VLMs) like GPT-4V and Vicuna~\citep{DBLP:journals/corr/abs-2304-03277}.
We have also extended Motion-X with CoT. And we label the two models, the one with CoT and the one without CoT, as $\text{H-GPT++}_\text{w.o. CoT}$ and $\text{H-GPT++}$ respectively.
Since Motion-X includes more and more diverse motion sequences, we retrained the motion tokenizer with different settings and found that the reconstruction performed best when the codebook number was $1024$ and the codebook dimension was $2048$. Therefore, we adopted this configuration in our experiments. The remaining settings are largely consistent with the training parameters of HumanML3D-X.

The evaluation results are shown in Table~\ref{tab:hgpt_motionx}. We observed that, different from the results reported in~\citet{DBLP:conf/nips/LinZLCZWZ23}, training on Motion-X can achieve decent performance, but the overall performance is not as good as directly using the training data from HumanML3D-X.
We guess that part of the reason may be due to the fact that the text labeled by VLMs and human motion data estimated by Motion-X from videos still exhibits certain differences in distribution and quality compared to the precise human motion capture data with human labeled language descriptions like HumanML3D-X.
For example, in Motion-X, a certain proportion of motion textual labels are phrased like \textit{``the subject of the sentence is `a man.'''} and \textit{``the existence of life is a fact.''}
Meanwhile, we can observe that the overall performance of the model has improved after incorporating CoT.
It appears that there is still considerable room for improvement in the research community's approach of leveraging scaling human pose data estimated from videos to enhance the overall generalization capability of models.
%
% We have placed more details about the Motion-X experiment in Appendix~\ref{app:hgpt_motionx}.

\paragraph{CoT Generation Performance} To further evaluate the generalization ability of the CoT version model for complex and abstract instructions, we designed two categories comprising 50 test instruction cases to carefully compare the performance of models without CoT and those with CoT. 
Complex instructions are constructed by randomly and naturally combining two original instructions from HumanML3D-X using GPT-4o. For example: \textit{``A person is running in place, then starts to run right and left, while jumping up and down.''}
Abstract instructions involve randomly generating 10 examples each for imitative performance, actions with emotion, interaction with human, interaction with objects, and interaction with the environment using GPT-4o. For instance: \textit{``Pantomime pulling a heavy rope hand over hand, showing the strain in your arms and back.''}

Since there is no ground truth motion for these instructions, making it inconvenient to establish evaluation metrics, we adopt the approach of manually observing the generated motions and comparing them to assess these models.
We did not assign separate scores to the results generated by each model. Instead, we determined which model produced better results—including cases where they perform equally well—through comparative evaluation, and then calculated the final scores.
In the evaluation task of 50 complex instructions, the score between $\text{H-GPT}_\text{w.o. CoT}$ and $\text{H-GPT}$ is $23:27$, and the score among $\text{H-GPT}_\text{w.o. CoT}$, $\text{H-GPT}$, $\text{H-GPT++}_\text{w.o. CoT}$, and $\text{H-GPT++}$ is $9:18:12:28$.
In the evaluation task of 50 abstract instructions, the score between $\text{H-GPT}_\text{w.o. CoT}$ and $\text{H-GPT}$ is $23:30$, and the score among $\text{H-GPT}_\text{w.o. CoT}$, $\text{H-GPT}$, $\text{H-GPT++}_\text{w.o. CoT}$, and $\text{H-GPT++}$ is $17:26:12:20$.
As we can see, through CoT enhancement, both the H-GPT and H-GPT++ models have achieved better generation results. However, no consistent superiority between these two models has been observed yet.
We visualize several typical examples in Figure~\ref{fig:hgpt_cot}. For more test cases of CoT, please refer to Appendix~\ref{app:hgpt_cot_eval}.

% \newpage
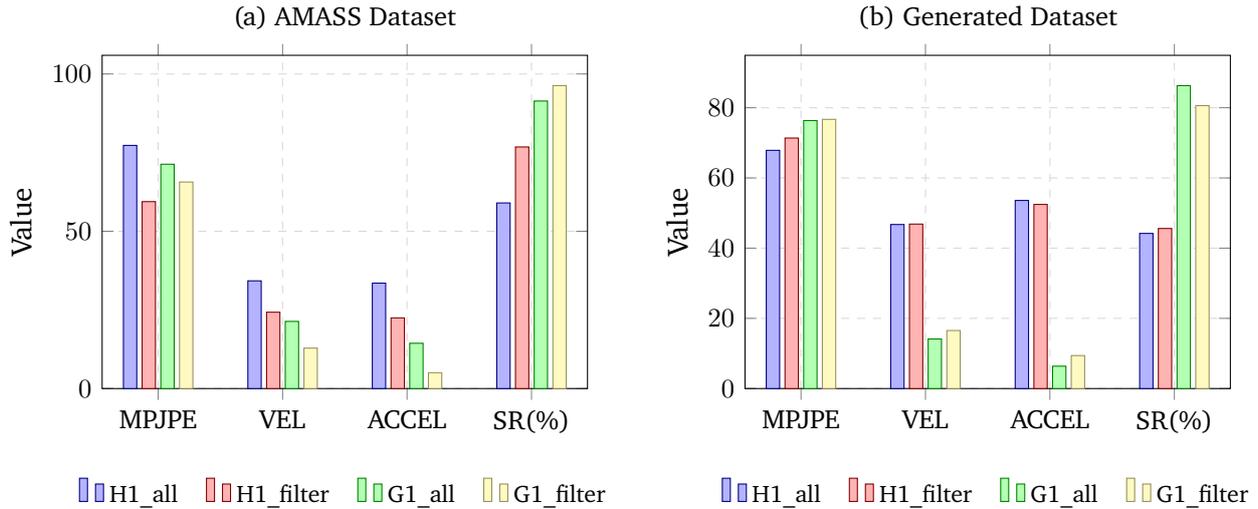
\begin{figure}[tbp]
    \centering
    \begin{subfigure}[b]{0.48\textwidth}
        \centering
        \begin{tikzpicture}
            \begin{axis}[
                ybar,
                bar width=5pt,
                enlarge x limits=0.15,
                legend style={
                    at={(0.5,-0.25)},
                    anchor=north,
                    legend columns=4,
                    draw=none,
                    /tikz/every even column/.append style={column sep=0.3cm},
                    font=\footnotesize
                },
                title={(a) AMASS Dataset},
                ylabel={Value},
                symbolic x coords={MPJPE, VEL, ACCEL, SR(\%)},
                xtick=data,
                ymin=0,
                height=6cm,
                width=\textwidth,
                grid=major,
                grid style={dashed,gray!30},
                tick label style={font=\small},
                title style={font=\small}
            ]

            % H1_all
            \addplot+[
                style={fill=blue!30,draw=blue!50!black}
            ] coordinates {
                (MPJPE, 77.29)
                (VEL, 34.238)
                (ACCEL, 33.514)
                (SR(\%), 59.0)
            };
            \addlegendentry{H1\_all}

            % H1_filter
            \addplot+[
                style={fill=red!30,draw=red!50!black}
            ] coordinates {
                (MPJPE, 59.424)
                (VEL, 24.27)
                (ACCEL, 22.468)
                (SR(\%), 76.8)
            };
            \addlegendentry{H1\_filter}

            % G1_all
            \addplot+[
                style={fill=green!30,draw=green!50!black}
            ] coordinates {
                (MPJPE, 71.306)
                (VEL, 21.348)
                (ACCEL, 14.396)
                (SR(\%), 91.4)
            };
            \addlegendentry{G1\_all}

            % G1_filter
            \addplot+[
                style={fill=yellow!30,draw=yellow!50!black}
            ] coordinates {
                (MPJPE, 65.646)
                (VEL, 12.854)
                (ACCEL, 5.04)
                (SR(\%), 96.3)
            };
            \addlegendentry{G1\_filter}

            \end{axis}
        \end{tikzpicture}
    \end{subfigure}
    \hfill
    \begin{subfigure}[b]{0.48\textwidth}
        \centering
        \begin{tikzpicture}
            \begin{axis}[
                ybar,
                bar width=5pt,
                enlarge x limits=0.15,
                legend style={
                    at={(0.5,-0.25)},
                    anchor=north,
                    legend columns=4,
                    draw=none,
                    /tikz/every even column/.append style={column sep=0.3cm},
                    font=\footnotesize
                },
                title={(b) Generated Dataset},
                ylabel={Value},
                symbolic x coords={MPJPE, VEL, ACCEL, SR(\%)},
                xtick=data,
                ymin=0,
                height=6cm,
                width=\textwidth,
                grid=major,
                grid style={dashed,gray!30},
                tick label style={font=\small},
                title style={font=\small}
            ]

            % H1_all
            \addplot+[
                style={fill=blue!30,draw=blue!50!black}
            ] coordinates {
                (MPJPE, 67.866)
                (VEL, 46.734)
                (ACCEL, 53.586)
                (SR(\%), 44.2)
            };
            \addlegendentry{H1\_all}

            % H1_filter
            \addplot+[
                style={fill=red!30,draw=red!50!black}
            ] coordinates {
                (MPJPE, 71.368)
                (VEL, 46.858)
                (ACCEL, 52.456)
                (SR(\%), 45.6)
            };
            \addlegendentry{H1\_filter}

            % G1_all
            \addplot+[
                style={fill=green!30,draw=green!50!black}
            ] coordinates {
                (MPJPE, 76.342)
                (VEL, 14.128)
                (ACCEL, 6.374)
                (SR(\%), 86.3)
            };
            \addlegendentry{G1\_all}

            % G1_filter
            \addplot+[
                style={fill=yellow!30,draw=yellow!50!black}
            ] coordinates {
                (MPJPE, 76.689)
                (VEL, 16.530)
                (ACCEL, 9.396)
                (SR(\%), 80.6)
            };
            \addlegendentry{G1\_filter}

            \end{axis}
        \end{tikzpicture}
    \end{subfigure}
    \caption{Comparison of different metrics for evaluating on HumanML3D-X dataset and \texttt{H-GPT} generated dataset.  MPJPE values are divided by 3, VEL and ACCEL values are multiplied by 2.}
    \label{fig:hact_rpt_results}
\end{figure}

\begin{figure}[tb]
    \centering
    \includegraphics[width=0.9\textwidth]{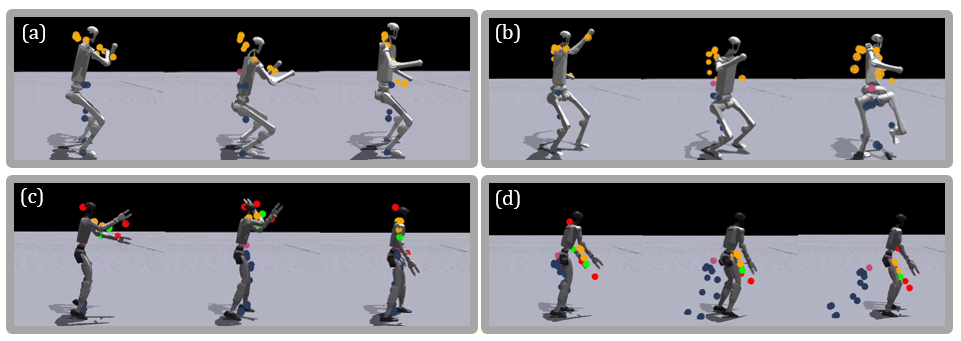}
    \caption{H-ACT RPT Simulation Tracking Results. (a) Prompt: \textit{``Shiver and hug yourself tightly to communicate feeling cold or scared.''} (b) Prompt: \textit{``Do a simple, three-step robot dance to a funky beat you imagine.''} (c) Prompt: \textit{``A person frantically digging through the dirt with bare hands, nails cracking, driven by desperate hope.''} (d) Prompt: \textit{``A person stumbling and catching themselves on a lamppost, swaying slightly before righting their posture with false dignity.''} (a) and (c) are examples of successful tracking for the Unitree H1 and G1 robots, while (b) and (d) are examples of tracking failures for the Unitree H1 and G1 robots.
}
    \label{fig:hact_rpt_vis}
\end{figure}

\subsection{Humanoid Whole-Body Control in Simulation Environments}
To answer \textbf{Q2}, we conducted a series of experimental validations of the \texttt{H-ACT} module using the H1 and G1 humanoid robots in the IsaacGym~\cite{DBLP:conf/nips/MakoviychukWGLS21} and MuJoCo~\citep{DBLP:conf/iros/TodorovET12} physics simulators.
We focus on the performance of reinforcement learning (RL) based control during the pre-training stage, its tracking effect on motions generated by \texttt{H-GPT}, and the performance of our proposed RL fine-tuning method in the inference stage.

\paragraph{RL Pre-Training (RPT)} To provide a good initialization parameter for the humanoid robot whole-body control policy, we first pre-trained a general motion controller using the AMASS human motion dataset~\citep{DBLP:conf/iccv/MahmoodGTPB19} with RL. 

% train
\textbf{Training} We first retarget the human motion data from AMASS to the skeleton structures of Unitree's H1 and G1 robots. 
Since the hands are typically lightweight and do not significantly affect balance, but have too many degrees of freedom that can reduce the efficiency of RL training, we did not add hand joints to these humanoid robots in the RL framework. Therefore, here we map the 24-joint SMPL 3D human motions to the 19-DOF H1 and 21-DOF G1 robot structures, respectively.
We trained these robots in NVIDIA's GPU-accelerated IsaacGym physics simulation to accurately follow the retargeted reference motions while maintaining body stability and preventing falls.
In terms of RL rewards, we set basic torque and DOF position limits, and applied regularization to situations such as excessively high DOF acceleration and velocity, prolonged feet air time, and stumbling. We penalized substantial differences between the actual motion and the reference motion in terms of joint positions, velocities, accelerations, body rotation angles, and speeds. Additionally, we imposed penalties for the humanoid robot falling, specifically when its center of mass fell below a set threshold.
For each robot, we launched 4096 parallel environments and optimized the policy using the PPO algorithm implemented in the RSL-RL framework~\citep{DBLP:journals/corr/abs-2509-10771}.

\textbf{Results} We focused on evaluating the MPJPE, VEL, and ACCEL metrics for the whole-body control policies of H1 and G1, which measure the differences in joint angles, velocities, and accelerations between the robot's actual states and the reference motions. We compared the performance of policies trained using the teacher-student strategy with both raw and filtered data. The results are shown in Figure~\ref{fig:hact_rpt_results} (a).
From the figure, it can be observed that using the filtered data consistently improves all metrics of both robot policies, both on the complete AMASS data and on the filtered AMASS data.
Additionally, it can be observed that all metrics of H1 (SR 40\% $\sim$ 50\%) are relatively poorer compared to those of G1 (SR 80\% $\sim$ 90\%). This, to some extent, indicates that the structural design of G1 enables it to perform diverse actions more stably and easily than the H1 robot.
Furthermore, we found that factors such as the stability of the reference's initial pose and the rate of change in body and joint motions significantly influence the success rate of policy tracking.

To further evaluate the performance of the pretrained controllers of H1 and G1, we tested them using the out-of-domain complex and abstract test samples generated by \texttt{H-GPT} from the previous subsection, with the results shown in Figure ~\ref{fig:hact_rpt_results}(b). 
From this, it can be seen that the tracking performance of H1 and G1 on these newly generated motions shows a more significant difference, and the model trained on filtered data does not exhibit better performance. We speculate that this is partly because the generated actions are noisier, and training on data with noise may enhance the model's robustness.
The visualization results of some successful and failed cases are shown in Figure~\ref{fig:hact_rpt_vis}.
%
% The retargeting details can be found in Appendix~\ref{app:hact_retarget}. And for more details and results regarding RL pre-training, please refer to the Appendix~\ref{app:hact_rpt}.

\paragraph{RL Fine-tuning (RFT)} To further enhance the motion tracking performance of the controller during inference, we propose a strategy that continues to leverage physics simulation for reinforcement learning-based imitation learning at this stage. 
Specifically, during inference, we initialize the controller with pre-trained checkpoints and employ a teacher-student framework to conduct targeted training tailored to the reference motion.
This way we can iterate over a small number of steps in a short time to make the controller better fit the reference motion to be tracked.

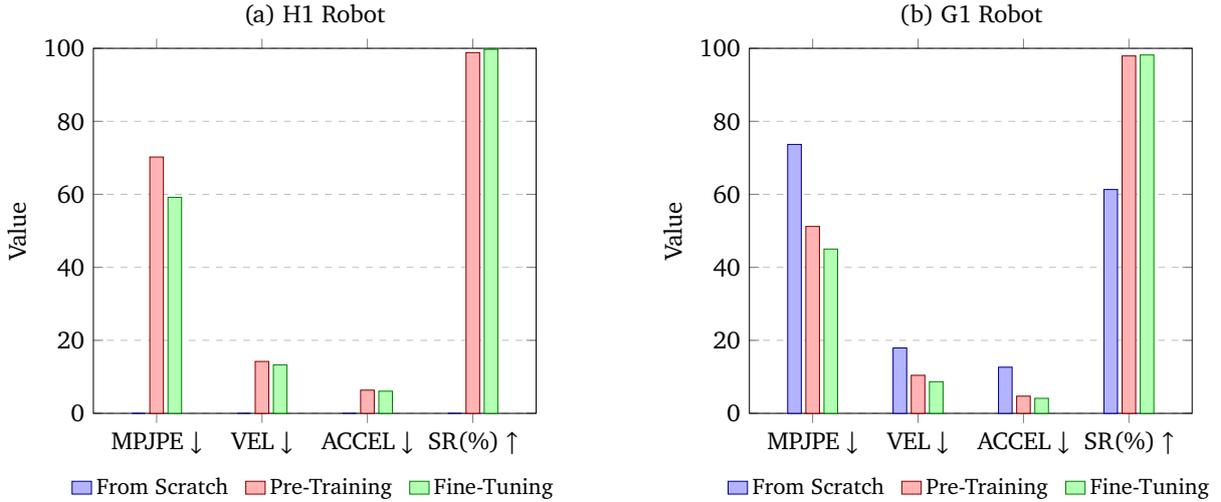
\begin{figure}[tbp]
    \centering
    \begin{subfigure}{0.48\textwidth}
        \centering
        \begin{tikzpicture}[scale=0.85]
        \begin{axis}[
            title={(a) H1 Robot},
            ybar,
            bar width=6pt,
            ymin=0,
            ymax=100,
            ytick={0,20,40,60,80,100},
            yticklabels={0,20,40,60,80,100},
            xtick={0,1,2,3},
            xticklabels={
                $\text{MPJPE}\downarrow$,
                $\text{VEL}\downarrow$,
                $\text{ACCEL}\downarrow$,
                $\text{SR(\%)}\uparrow$
            },
            xticklabel style={align=center},
            xlabel={},
            ylabel={Value},
            % 将图例移动到图下方
            legend style={
                at={(0.5,-0.15)},
                anchor=north,
                legend columns=3,
                font=\small,
                draw=none,
                fill=none,
                cells={anchor=center},
                /tikz/every even column/.append style={column sep=0.5em}
            },
            enlarge x limits=0.2,
            ymajorgrids=true,
            grid style=dashed,
            cycle list={
                {fill=blue!30,draw=blue!50!black},
                {fill=red!30,draw=red!50!black},
                {fill=green!30,draw=green!50!black}
            },
            legend image code/.code={%
                \draw[#1] (0cm,-0.1cm) rectangle (0.3cm,0.1cm);
            }
        ]
        
        % From Scratch Training - H1
        \addplot+ coordinates {
            (0,0)        % MPJPE (no data)
            (1,0)        % VEL (no data)
            (2,0)        % ACCEL (no data)
            (3,0)        % SR
        };
        
        % Pre-Training - H1 (MPJPE divided by 2: 140.439/2 = 70.2195)
        % VEL multiplied by 2: 7.11014*2 = 14.22028
        % ACCEL multiplied by 2: 3.17803*2 = 6.35606
        \addplot+ coordinates {
            (0,70.2195)   % MPJPE (140.439/2)
            (1,14.22028)  % VEL (7.11014*2)
            (2,6.35606)   % ACCEL (3.17803*2)
            (3,98.835)    % SR (0.98835*100)
        };
        
        % Fine-Tuning - H1 (MPJPE divided by 2: 118.362/2 = 59.181)
        % VEL multiplied by 2: 6.64672*2 = 13.29344
        % ACCEL multiplied by 2: 3.04014*2 = 6.08028
        \addplot+ coordinates {
            (0,59.181)    % MPJPE (118.362/2)
            (1,13.29344)  % VEL (6.64672*2)
            (2,6.08028)   % ACCEL (3.04014*2)
            (3,99.734)    % SR (0.99734*100)
        };
        
        \legend{From Scratch, Pre-Training, Fine-Tuning}
        
        \end{axis}
        \end{tikzpicture}
    \end{subfigure}
    \hfill
    \begin{subfigure}{0.48\textwidth}
        \centering
        \begin{tikzpicture}[scale=0.85]
        \begin{axis}[
            title={(b) G1 Robot},
            ybar,
            bar width=6pt,
            ymin=0,
            ymax=100,
            ytick={0,20,40,60,80,100},
            yticklabels={0,20,40,60,80,100},
            xtick={0,1,2,3},
            xticklabels={
                $\text{MPJPE}\downarrow$,
                $\text{VEL}\downarrow$,
                $\text{ACCEL}\downarrow$,
                $\text{SR(\%)}\uparrow$
            },
            xticklabel style={align=center},
            xlabel={},
            ylabel={Value},
            % 将图例移动到图下方
            legend style={
                at={(0.5,-0.15)},
                anchor=north,
                legend columns=3,
                font=\small,
                draw=none,
                fill=none,
                cells={anchor=center},
                /tikz/every even column/.append style={column sep=0.5em}
            },
            enlarge x limits=0.2,
            ymajorgrids=true,
            grid style=dashed,
            cycle list={
                {fill=blue!30,draw=blue!50!black},
                {fill=red!30,draw=red!50!black},
                {fill=green!30,draw=green!50!black}
            },
            legend image code/.code={%
                \draw[#1] (0cm,-0.1cm) rectangle (0.3cm,0.1cm);
            }
        ]
        
        % From Scratch Training - G1 (MPJPE divided by 2: 147.3577/2 = 73.67885)
        % VEL multiplied by 2: 8.96003*2 = 17.92006
        % ACCEL multiplied by 2: 6.3216*2 = 12.6432
        \addplot+ coordinates {
            (0,73.67885)  % MPJPE (147.3577/2)
            (1,17.92006)  % VEL (8.96003*2)
            (2,12.6432)   % ACCEL (6.3216*2)
            (3,61.3279947) % SR (0.613279947*100)
        };
        
        % Pre-Training - G1 (MPJPE divided by 2: 102.402/2 = 51.201)
        % VEL multiplied by 2: 5.21187*2 = 10.42374
        % ACCEL multiplied by 2: 2.3566*2 = 4.7132
        \addplot+ coordinates {
            (0,51.201)    % MPJPE (102.402/2)
            (1,10.42374)  % VEL (5.21187*2)
            (2,4.7132)    % ACCEL (2.3566*2)
            (3,97.918)    % SR (0.97918*100)
        };
        
        % Fine-Tuning - G1 (MPJPE divided by 2: 90.0223/2 = 45.01115)
        % VEL multiplied by 2: 4.33317*2 = 8.66634
        % ACCEL multiplied by 2: 2.05893*2 = 4.11786
        \addplot+ coordinates {
            (0,45.01115)  % MPJPE (90.0223/2)
            (1,8.66634)   % VEL (4.33317*2)
            (2,4.11786)   % ACCEL (2.05893*2)
            (3,98.185)    % SR (0.98185*100)
        };
        
        \legend{From Scratch, Pre-Training, Fine-Tuning}
        
        \end{axis}
        \end{tikzpicture}
    \end{subfigure}
    \caption{Performance comparison of different training strategies on H1 and G1 robots. MPJPE values are divided by 2 for better visualization. VEL and ACCEL values are multiplied by 2 for better visualization. SR values are shown as percentages.}
    \label{fig:hact_rft}
\end{figure}

We randomly sampled 30 motions from the AMASS dataset to test the control policies for H1 and G1. We primarily compared the effectiveness of this fine-tuning strategy against training from scratch and using pre-training alone. The main evaluation metrics were MPJPE and Success Rate. 
As shown in Figure ~\ref{fig:hact_rft}, we can see that the MPJPE and SR metrics for motion tracking have improved steadily after RFT on both the H1 and G1 robots. We also observed that some previously less stable tracking motions became more stable after RFT.

\begin{figure}[tbp]
    \centering
    \begin{subfigure}[b]{0.48\linewidth}
        \centering
        \begin{tikzpicture}[scale=0.85]
            \begin{axis}[
            title={(a) H1 Robot},
                width=\linewidth,
                height=6cm,
                xlabel={Steps},
                ylabel={Metric Value},
                ymin=0,
                ymax=100,
                xtick={0,500,1000,1500,2000},
                grid=both,
                legend style={at={(0.98,1.0)},anchor=north east},
                legend cell align={left},
                legend style={font=\footnotesize},
                x tick label style={rotate=0,font=\small},
                y tick label style={/pgf/number format/fixed},
                cycle list name=color list,
                title style={yshift=-5pt}
            ]
            
            % Success Rate (SR) * 100
            \addplot+[thick, mark=*, mark size=1.5pt] 
                coordinates {
                    (0, 0.975586 * 100)
                    (250, 0.997070313 * 100)
                    (500, 0.997395833 * 100)
                    (750, 0.997558594 * 100)
                    (1000, 0.998046875 * 100)
                    (1250, 0.998046875 * 100)
                    (1500, 0.997884115 * 100)
                    (1750, 0.997721354 * 100)
                    (2000, 0.998209635 * 100)
                };
            \addlegendentry{SR(\%) $\uparrow$}
            
            % MPJPE/2
            \addplot+[thick, mark=square*, mark size=1.3pt] 
                coordinates {
                    (0, 150.6705/2)
                    (250, 125.2686667/2)
                    (500, 120.8453333/2)
                    (750, 116.9213333/2)
                    (1000, 116.5741667/2)
                    (1250, 114.0866667/2)
                    (1500, 115.5345/2)
                    (1750, 113.069/2)
                    (2000, 115.5328333/2)
                };
            \addlegendentry{MPJPE $\downarrow$}
            
            % Velocity (VEL) * 2
            \addplot+[thick, mark=triangle*, mark size=1.5pt] 
                coordinates {
                    (0, 7.421833 * 2)
                    (250, 7.227666667 * 2)
                    (500, 7.187166667 * 2)
                    (750, 7.140666667 * 2)
                    (1000, 7.130166667 * 2)
                    (1250, 7.1405 * 2)
                    (1500, 7.141166667 * 2)
                    (1750, 7.185666667 * 2)
                    (2000, 7.181 * 2)
                };
            \addlegendentry{VEL $\downarrow$}
            
            % Acceleration (ACCEL) * 2
            \addplot+[thick, mark=diamond*, mark size=1.5pt] 
                coordinates {
                    (0, 3.305333 * 2)
                    (250, 3.315166667 * 2)
                    (500, 3.342833333 * 2)
                    (750, 3.294333333 * 2)
                    (1000, 3.292 * 2)
                    (1250, 3.294666667 * 2)
                    (1500, 3.345833333 * 2)
                    (1750, 3.366 * 2)
                    (2000, 3.351666667 * 2)
                };
            \addlegendentry{ACCEL $\downarrow$}
            
            \end{axis}
        \end{tikzpicture}
    \end{subfigure}
    \hfill
    \begin{subfigure}[b]{0.48\linewidth}
        \centering
        \begin{tikzpicture}[scale=0.85]
            \begin{axis}[
                title={(b) G1 Robot},
                width=\linewidth,
                height=6cm,
                xlabel={Steps},
                ylabel={Metric Value},
                ymin=0,
                ymax=100,
                xtick={0,500,1000,1500,2000},
                grid=both,
                legend style={at={(0.98,0.92)},anchor=north east},
                legend cell align={left},
                legend style={font=\footnotesize},
                x tick label style={rotate=0,font=\small},
                y tick label style={/pgf/number format/fixed},
                cycle list name=color list,
                title style={yshift=-5pt}
            ]
            
            % Success Rate (SR) * 100 - 从表格中获取
            \addplot+[thick, mark=*, mark size=1.5pt] 
                coordinates {
                    (0, 0.971 * 100)
                    (250, 0.969 * 100)
                    (500, 0.963 * 100)
                    (750, 0.975 * 100)
                    (1000, 0.971 * 100)
                    (1250, 0.969 * 100)
                    (1500, 0.967 * 100)
                    (1750, 0.972 * 100)
                    (2000, 0.971 * 100)
                };
            \addlegendentry{SR(\%) $\uparrow$}
            
            % MPJPE/2 - mpipe_g列的值除以2
            \addplot+[thick, mark=square*, mark size=1.3pt] 
                coordinates {
                    (0, 103.466/2)
                    (250, 98.497/2)
                    (500, 90.275/2)
                    (750, 92.446/2)
                    (1000, 91.031/2)
                    (1250, 91.764/2)
                    (1500, 90.287/2)
                    (1750, 92.884/2)
                    (2000, 91.506/2)
                };
            \addlegendentry{MPJPE $\downarrow$}
            
            % Velocity (VEL) * 2 - vel_dist列的值乘以2
            \addplot+[thick, mark=triangle*, mark size=1.5pt] 
                coordinates {
                    (0, 5.132 * 2)
                    (250, 4.808 * 2)
                    (500, 4.691 * 2)
                    (750, 4.512 * 2)
                    (1000, 4.388 * 2)
                    (1250, 4.331 * 2)
                    (1500, 4.258 * 2)
                    (1750, 4.287 * 2)
                    (2000, 4.146 * 2)
                };
            \addlegendentry{VEL $\downarrow$}
            
            % Acceleration (ACCEL) * 2 - accel_dist列的值乘以2
            \addplot+[thick, mark=diamond*, mark size=1.5pt] 
                coordinates {
                    (0, 2.373 * 2)
                    (250, 2.292 * 2)
                    (500, 2.373 * 2)
                    (750, 2.265 * 2)
                    (1000, 2.15 * 2)
                    (1250, 2.107 * 2)
                    (1500, 2.132 * 2)
                    (1750, 2.136 * 2)
                    (2000, 2.061 * 2)
                };
            \addlegendentry{ACCEL $\downarrow$}
            
            \end{axis}
        \end{tikzpicture}
    \end{subfigure}
    \caption{Performance metrics at different training steps. MPJPE values are divided by 2 for better visualization.}
    \label{fig:hact_rft_scaling}
\end{figure}
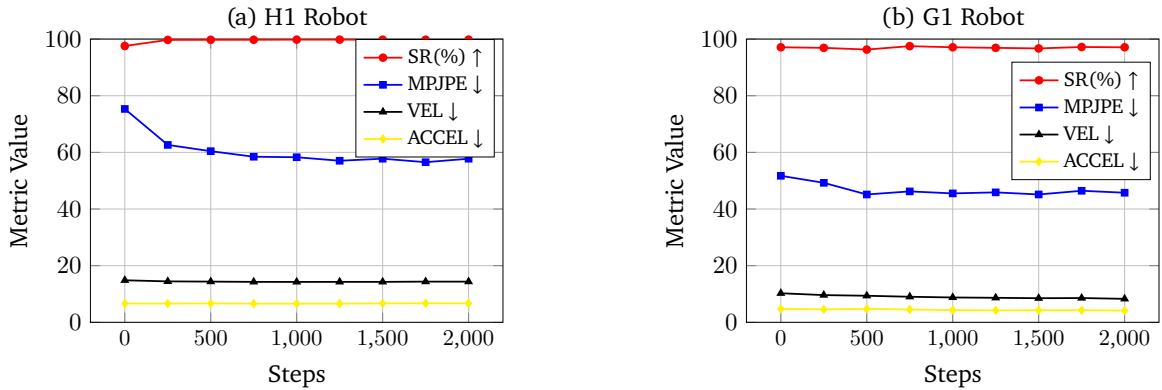

\begin{figure}[tb]
    \centering
    \includegraphics[width=0.9\textwidth]{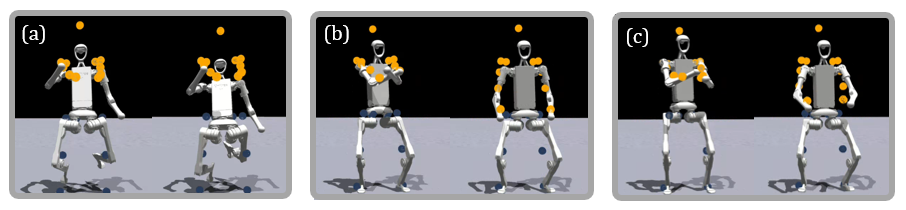}
    \caption{H-ACT RFT Simulation Tracking Results. Prompt: \textit{``Shiver and hug yourself tightly to communicate feeling cold or scared.''} (a) Training from scratch. (b) Pre-train. (c) Fine-tune}
    \label{fig:hact_rft_vis}
\end{figure}

We further tested the variation of several metrics with the number of tuning steps during the RL Fine-tuning phase, and the results are shown in Figure~\ref{fig:hact_rft_scaling}. From subplots (a) and (b) of the figure, it can be seen that the model has already achieved a significant improvement around 500 steps. A visual comparison case of training from scratch, pretraining, and fine-tuning is shown in Figure~\ref{fig:hact_rft_vis}. 
% We have placed more implementation details and results in the Appendix~\ref{app:hact_rft}.

\begin{figure}[tb]
    \centering
    \begin{subfigure}[t]{0.49\textwidth}
        \centering
        \includegraphics[width=\linewidth]{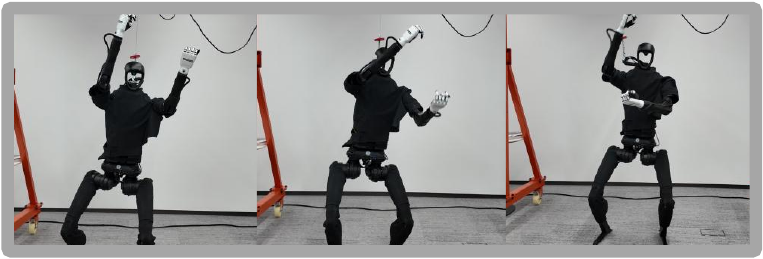}
        \caption{Unitree H1 Robot. Prompt: \textit{``A person is playing the role of an elephant.''}}
        \label{fig:hact_rft_a1}
    \end{subfigure}
    \hfill
    \begin{subfigure}[t]{0.49\textwidth}
        \centering
        \includegraphics[width=\linewidth]{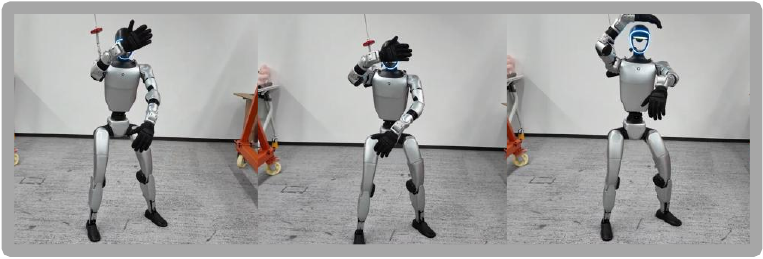}
        \caption{Unitree G1 Robot. Prompt: \textit{``A person is playing the role of an elephant.''}}
        \label{fig:hact_rft_b1}
    \end{subfigure}

    \medskip
    \begin{subfigure}[t]{0.49\textwidth}
        \centering
        \includegraphics[width=\linewidth]{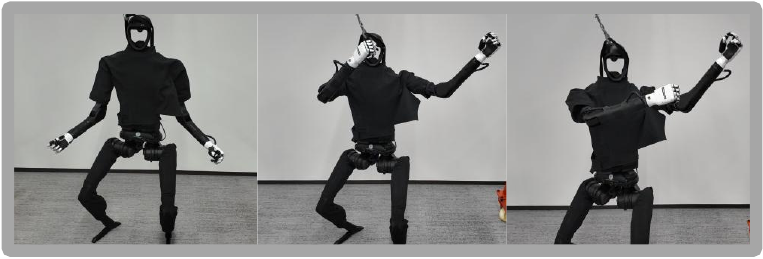}
        \caption{Unitree H1 Robot. Prompt: \textit{``A person is playing a violin.''}}
        \label{fig:hact_rft_c1}
    \end{subfigure}
    \hfill
    \begin{subfigure}[t]{0.49\textwidth}
        \centering
        \includegraphics[width=\linewidth]{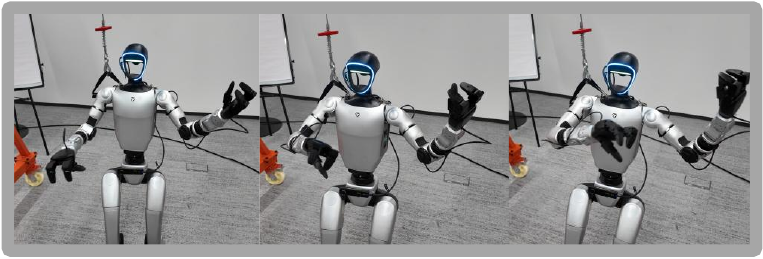}
        \caption{Unitree G1 Robot. Prompt: \textit{``A person is playing a violin.''}}
        \label{fig:hact_rft_a2}
    \end{subfigure}

    \medskip
    \begin{subfigure}[t]{0.49\textwidth}
        \centering
        \includegraphics[width=\linewidth]{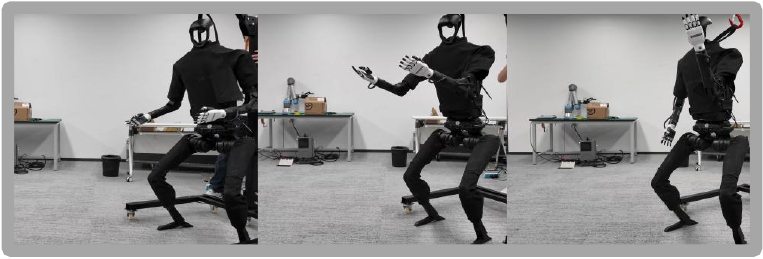}
        \caption{Unitree H1 Robot. Prompt: \textit{``A person took a few steps forward and then answered a phone call.''}}
        \label{fig:hact_rft_b2}
    \end{subfigure}
    \hfill
    \begin{subfigure}[t]{0.49\textwidth}
        \centering
        \includegraphics[width=\linewidth]{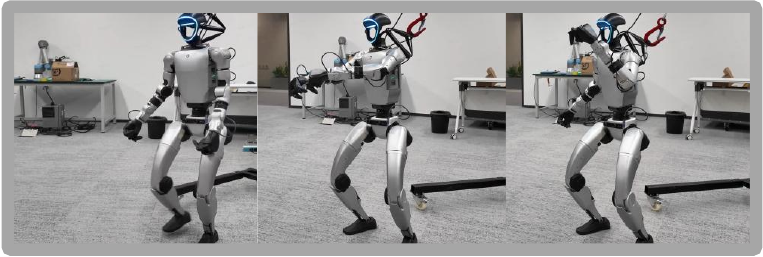}
        \caption{Unitree G1 Robot. Prompt: \textit{``A person took a few steps forward and then answered a phone call.''}}
        \label{fig:hact_rft_c2}
    \end{subfigure}

    \caption{H-ACT Real-World Depolyment Results. In the figure, (a) and (b) illustrate the scenario where the H1 robot and the G1 robot complete abstract instructions. (c) and (d) demonstrate cases involving typical hand movements, while (e) and (f) depict situations of moving first and then imitating actions.}
    \label{fig:hact_real}
\end{figure}

\begin{figure}[htb]
    \centering
    % 第一排
    \begin{subfigure}[t]{0.49\textwidth}
        \centering
        \includegraphics[width=\linewidth]{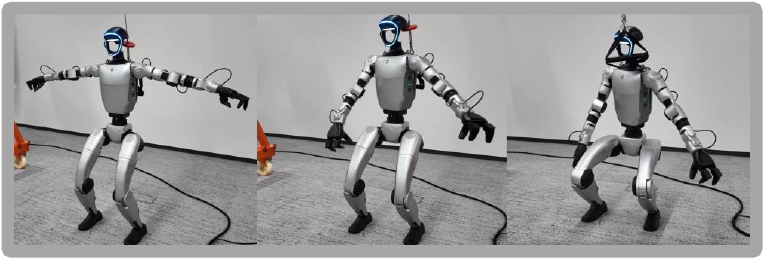}
        \caption{The HugWBC Policy. Prompt: \textit{``A person is practicing squats.''}}
        \label{fig:hact_rft_a1}
    \end{subfigure}
    \hfill
    \begin{subfigure}[t]{0.49\textwidth}
        \centering
        \includegraphics[width=\linewidth]{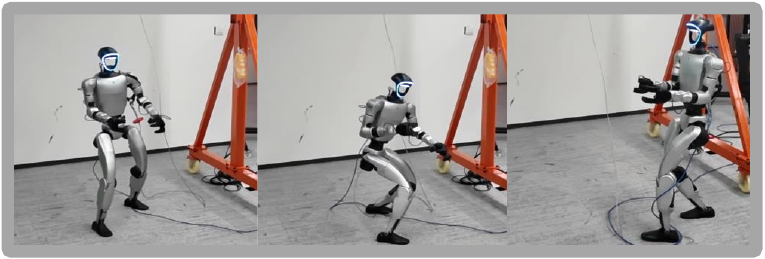}
        \caption{The TWIST Policy. Prompt: \textit{``A person dancing alone in a messy kitchen, swaying with a broom as a partner to a silent radio tune.''}}
        \label{fig:hact_rft_b1}
    \end{subfigure}
    \caption{H-ACT Cross-Policy Depolyment Results. The \texttt{H-ACT} framework can efficiently support the stable deployment of different control strategies.}
    \label{fig:hact_otherpolicy}
\end{figure}

\begin{figure}[htb]
    \centering
    \begin{subfigure}[t]{0.49\textwidth}
        \centering
        \includegraphics[width=\linewidth]{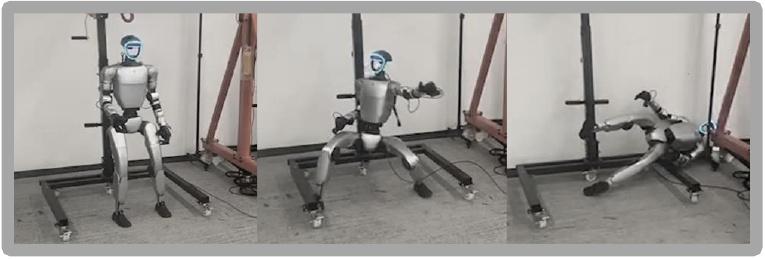}
        \caption{The TWIST Policy. Prompt: \textit{``A person crouching low to the ground, body coiled like a spring, ready to launch into action.''}}
        \label{fig:hact_rft_a1}
    \end{subfigure}
    \hfill
    \begin{subfigure}[t]{0.49\textwidth}
        \centering
        \includegraphics[width=\linewidth]{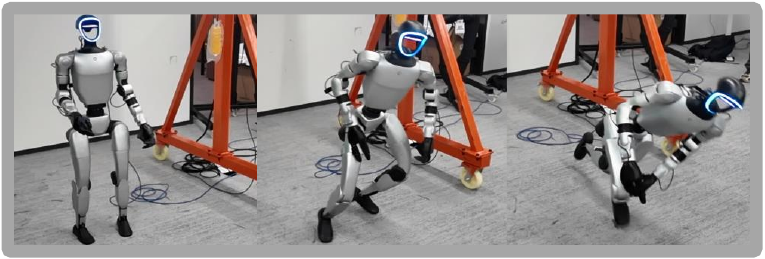}
        \caption{The TWIST Policy. Prompt: \textit{``A person sprinting through the rain, their arms pumping wildly as if trying to outrun a storm of memories.''}} 
        \label{fig:hact_rft_b1}
    \end{subfigure}
    \caption{H-ACT Real-World Failure Cases. (a) shows the Unitree G1 robot falling over during an extreme squatting motion, and (b) shows the robot failing to track a fast forward-moving target motion.}
    \label{fig:hact_badcases}
\end{figure}

\subsection{Performance of the Entire Framework in the Real World}  

Finally, to address \textbf{Q3}, we fully deployed the entire \texttt{FRoM-W1} framework on the Unitree H1 and G1 robots.
In addition to testing with the default RL Policy of \texttt{H-ACT}, as mentioned in the previous sections, our framework is compatible with many RL control algorithms in the field of motion mimicing. We developed a lightweight deployment framework for efficiently deploying these different policies. And we thus tested the execution capabilities of different RL Policies on the motions generated by \texttt{H-GPT}.

\paragraph{Setup} We first deployed our default policy on the H1 and G1 robots. We used a 4090 workstation as the edge-side computing resource to run the \texttt{H-GPT} model for motion generation. The generated motions were then retargeted and transmitted to the robots via wired or wireless connections. Finally, the whole-body motion controller of the \texttt{H-ACT} part was executed using the computing board deployed on the robots.
In addition to testing the default RL Policy of \texttt{H-ACT}, we also tested whole-body motion control policies for humanoid robots provided by HugWBC~\citep{xue2025unifiedgeneralhumanoidwholebody} and
TWIST~\citep{DBLP:journals/corr/abs-2505-02833}.
HugWBC is a unified whole-body controller that empowers humanoid robots to execute versatile gaits like jumping and hopping with real-time parameter tuning, while simultaneously accommodating upper-body interventions for loco-manipulation.
%
% BeyondMimic is a humanoid control framework that uses a diffusion model to learn agile motions from human demonstrations and can then flexibly recombine and adapt these skills in real-time to achieve complex, unseen goals, transferring zero-shot to real robots.
%
TWIST is a teleoperation system that enables humanoid robots to perform versatile whole-body skills through human motion imitation using a unified neural network controller.
Our hardware configuration and deployment framework can be referred to in Appendix~\ref{app:hact_real_setup} and~\ref{app:hact_robojudo}, respectively.

\paragraph{Results}
Typical examples of deploying the default reinforcement learning strategy after \texttt{H-GPT} generates the motion is shown in Figure~\ref{fig:hact_real}. 
These examples demonstrate several features of our overall framework, including the ability to comprehend and generate abstract instructions, the capacity to generate hand movements, and the stable deployment capability for locomotion actions.
Figure~\ref{fig:hact_otherpolicy} demonstrates the compatibility of our framework with different control strategies, enabling efficient and stable real-world deployment of models such as HugWBC and TWIST.
In addition to these successful examples, we found that the overall framework still faces issues in generating actions for arbitrary instructions. These problems include misalignment between generated actions and instructions, instability in the generated actions themselves, or scenarios where the actions cause collisions within the robot's body.
In addition, as shown in Figure~\ref{fig:hact_badcases}, for certain extreme movements—such as half-squats or rapid running—even if the robot can perform these actions in simulation, unstable conditions still frequently arise during real-world deployment due to the Sim2Real gap.
For more sim2real details, please refer to the Appendix~\ref{app:hact_real}.

% \newpage
\section{Related Work}

\paragraph{Text to Human Motion Generation} Text-conditioned 3D human motion generation is a fundamental task in 3D computer vision, which contributes to downstream applications such as VR content creation, gaming, and robotics.
In recent years, the introduction of the HumanML3D text-to-motion dataset and the T2M method~\cite{DBLP:conf/cvpr/GuoZZ0JL022} has laid a new foundation for this direction. 
This has led to autoregressive modeling-based approaches such as TM2T~\citep{DBLP:conf/eccv/GuoZWC22}, T2M-GPT~\citep{DBLP:journals/corr/abs-2301-06052}, and MotionGPT~\citep{DBLP:conf/nips/JiangCLYYC23}, diffusion model-based methods like MDM~\citep{DBLP:journals/corr/abs-2209-14916}, MLD~\citep{DBLP:conf/cvpr/ChenJLHFCY23}, and MotionDiffuse~\citep{DBLP:journals/pami/ZhangCPHGYL24}, as well as methods based on masked transformers, such as MoMask~\citep{DBLP:conf/cvpr/GuoMJW024}.
However, the HumanML3D dataset has two major issues. 
Firstly, its data scale is too small, making it difficult for models trained on it to achieve strong generalization capabilities. 
Secondly, it is based on the SMPL body representation~\citep{DBLP:journals/tog/LoperM0PB15}, which does not include hand modeling. Yet, for humanoid robots, the hands are crucial for the expressiveness of overall motions and for tasks such as manipulation and loco-manipulation.
To address the first issue, MotionMillion~\citep{DBLP:journals/corr/abs-2507-07095} expanded the scale of SMPL human motion sequences to the million level by using human pose estimation on large-scale videos from the internet. 
And Motion-R1~\citep{DBLP:journals/corr/abs-2506-10353} enhances generalization capability for linguistic instructions by incorporating a CoT approach.
Regarding the second issue, Motion-X~\citep{DBLP:conf/nips/LinZLCZWZ23} is currently the largest publicly accessible motion dataset that utilizes SMPL-X~\citep{DBLP:conf/cvpr/PavlakosCGBOTB19}, which includes hand modeling.
Work based on Motion-X includes HumanTOMATO~\citep{DBLP:conf/icml/LuCZLZ0S24}, MotionGPT-2~\citep{DBLP:journals/corr/abs-2410-21747}, MotionCraft~\citep{DBLP:conf/aaai/BianZJLZLX25}, and others. However, these efforts either do not involve further training on existing large language models~\citep{DBLP:conf/icml/LuCZLZ0S24, DBLP:conf/aaai/BianZJLZLX25}, are not open-sourced~\citep{DBLP:conf/icml/LuCZLZ0S24, DBLP:journals/corr/abs-2410-21747}, or utilize only a small portion of the dataset~\citep{DBLP:conf/aaai/BianZJLZLX25}.
For our motion generation component, we require an open-source, whole-body human motion generation model with hand modeling that is built upon existing LLMs and trained on massive data to achieve stronger language generalization capabilities. Therefore, based on the LLaMA model~\citep{DBLP:journals/corr/abs-2407-21783} and incorporating CoT techniques, we train \texttt{H-GPT} on HumanML3D-X and Motion-X to create a whole-body motion generation model represented in SMPL-X format, and open-source it.

\paragraph{Humanoid Whole-Body Motion Tracking}
The development of hardware platforms for humanoid robots, such as the Unitree H1 and G1, has laid the foundation for advancements in humanoid robot control algorithms in recent years, and a series of human-to-humanoid teleoperation and whole-body motion control algorithms for humanoid robots have emerged.
Early explorations involved using reinforcement learning techniques to enable humanoid robot platforms to accomplish specific tasks such as smooth locomotion~\citep{DBLP:journals/corr/abs-2410-11825}, performing backflips~\citep{DBLP:conf/humanoids/ChignoliKSK21}, walking on stepping stones~\citep{DBLP:journals/corr/abs-2502-10363}, and parkour~\citep{DBLP:conf/corl/ZhuangYZ24}.
Subsequently, researchers began exploring methods to enable humanoid robots to track human motions captured through motion capture systems, either offline or in real time, a process also known as motion mimicing or motion shadowing. 
ExBody~\citep{DBLP:conf/rss/ChengJCYYW24}, H2O~\citep{DBLP:conf/iros/HeLXZKLS24}, OmniH2O~\citep{DBLP:conf/corl/He0HXZ0KLS24}, and HumanPlus~\citep{DBLP:conf/corl/FuZWWF24} are several pioneering works that have explored in this direction. Among them, ExBody decouples and separately controls the upper and lower bodys of humanoid robots, while H2O, OmniH2O, and HumanPlus focus on whole-body control. Additionally, OmniH2O proposes a Teacher-Student training strategy, and HumanPlus further trains humanoid manipulation policies using data collected through teleoperation.
Subsequent works involve exploring more control methods, such as Vision Pros~\citep{DBLP:conf/icra/HeXLLXJKLSWFZ25,DBLP:journals/corr/abs-2511-02832}, motion capture devices~\citep{DBLP:journals/corr/abs-2505-02833}, exoskeleton cockpit systems~\citep{DBLP:journals/corr/abs-2502-13013}, diverse motion generation models~\citep{luo2025sonic}, and others,
as well as improving the model's effectiveness by enabling adaptive lower-body movements when the upper and lower bodys are decoupled~\citep{DBLP:journals/corr/abs-2505-03738}, further training on different subtasks separately~\citep{DBLP:journals/corr/abs-2412-13196}, and employing strategies such as the Mixture of Experts (MoE)~\citep{DBLP:journals/corr/abs-2506-14770} and merging expert policies~\citep{DBLP:journals/corr/abs-2508-08241}.
The purpose of the \texttt{H-ACT} module in \texttt{FRoM} is to have a universal whole-body tracker capable of mimicking human motions generated by \texttt{H-GPT}. Therefore, in principle, it can be compatible with the various whole-body control strategies mentioned above that can track human motion. Additionally, our universal deployment module supports the integration of the aforementioned multiple strategies~\citep{DBLP:conf/iros/HeLXZKLS24, DBLP:conf/corl/He0HXZ0KLS24, xue2025unifiedgeneralhumanoidwholebody,DBLP:journals/corr/abs-2505-02833, DBLP:journals/corr/abs-2511-02832,DBLP:journals/corr/abs-2508-08241} into our entire framework.
In our paper, we adopt a training framework similar to OmniH2O's teacher-student approach as our default strategy for whole-body motion controller training. Moreover, during the inference phase, an RL fine-tuning strategy is introduced to further enhance the tracker's accuracy and stability. This shares a similar concept with the third stage of the recent UniTracker~\citep{DBLP:journals/corr/abs-2507-07356}. However, to maintain simplicity and generality, we directly fine-tuned the entire network instead of adding a separate residual module.
% %
% Other tasks include tracking more agile movements~\citep{DBLP:journals/corr/abs-2502-01143, pan2025agilitymeet, DBLP:journals/corr/abs-2509-21690, xu2025learningagilestrikerskills}, introducing contact with objects~\citep{DBLP:conf/corl/ZhangXHS24, DBLP:journals/corr/abs-2502-01465, DBLP:conf/humanoids/BethalaHPAYWTF25, DBLP:journals/corr/abs-2505-06776,DBLP:journals/corr/abs-2509-26633,DBLP:journals/corr/abs-2510-05070,DBLP:journals/corr/abs-2510-14293,DBLP:journals/corr/abs-2510-17792,DBLP:journals/corr/abs-2510-26280}, and incorporating vision~\citep{DBLP:journals/corr/abs-2505-03729, DBLP:journals/corr/abs-2509-20322,DBLP:journals/corr/abs-2511-03996,he2025viralvisualsimtorealscale}.

\paragraph{Language-Drive Humanoid Whole-Body Control}
Whole-Body controlling humanoid robots using language, especially full-size robots such as the Unitree H1 and G1, remains a very new task.
RobotMDM~\citep{DBLP:conf/siggrapha/SerifiGK0B24} initially uses a small diffusion model to generate motions and trains a controller via RL techniques to control a very small robot in performing whole-body movements.
Harmon~\citep{DBLP:conf/corl/0002XLYZZ24} leverages off-the-shelf motion generation models to produce motion, and introduces hand and head movements by prompting visual language models.
UH-1~\citep{DBLP:journals/corr/abs-2412-14172} can be considered the first method to explore the use of vast video-based human motion data to achieve universal language control over the whole-body movements of full-size humanoid robots. However, it trains a small motion generation model from scratch without leveraging the capabilities of existing large models, and the expressiveness and diversity of the motions demonstrated by its control policy remain relatively limited. A more critical issue is that it trains a separate motion generation module for each humanoid robot, which is difficult to scale in scenarios involving massive data, high parameter models, and substantial computational consumption. In contrast, our approach chooses human motion models as a unified representation, allowing for natural transferability to humanoid robots with different structures.
Additionally, LangWBC~\citep{DBLP:journals/corr/abs-2504-21738} utilizes a small end-to-end network to achieve real-time language control of humanoid robot motions. LeVERB~\citep{DBLP:journals/corr/abs-2506-13751} employs its own designed small-scale two-brain network to control Unitree robots for full-body motion using latent verbs. Furthermore, the recent RoboGhost~\citep{DBLP:journals/corr/abs-2510-14952} also leverages latent command representations as direct conditions for generating the robot's final full-body action signals through a diffusion model.
In contrast, our entire framework aims to leverage and build upon the powerful language understanding capabilities and generalization potential of existing large models. Therefore, we have not designed or trained any small, standalone motion generation models from scratch.
RLPF~\citep{DBLP:journals/corr/abs-2506-12769} utilizes a large model to generate motion and then feeds back control-phase signals to the generation stage of the model to enhance its generation capabilities. However, it does not include hand movements, lacks CoT reasoning to improve model generalization, and has not open-sourced its model for use, evaluation, and comparison.
Concurrent work, Humanoid-LLA~\citep{liu2025commanding}, also employs CoT to enhance the model's generalization capability with language. It further integrates the vocabulary space of the large language model with the control signals of the humanoid robot, and then utilizes feedback from physical simulation to optimize the large model.
We introduced hand modeling, provided new benchmarks, scaled with more data, carefully evaluated the effects of incorporating CoT, optimized the RL training strategy, designed and implemented a universal sim2real module, and open-sourced the entire framework.
We hope to further advance and evaluate the boundaries of this entire field through this comprehensive framework, and promote the development of the field through open-source practices.

\section{Conclusion}
% \Doney{To Check}
% The experimental results demonstrate that \texttt{FRoM-W1} is a highly effective framework for controlling humanoid robots through natural language instructions, enabling them to showcase human skills. Its ability to generate accurate, stable, and generalizable motions, combined with successful real-world deployment, positions it as a significant advancement in humanoid robotics. 

In this work, we present \texttt{FRoM-W1}, an open-source framework that unifies language understanding, motion generation, and humanoid whole-body control. By combining \texttt{H-GPT} for versatile language-conditioned motion synthesis with \texttt{H-ACT} for cross-platform, policy-agnostic motion execution, \texttt{FRoM-W1} enables humanoid robots to follow natural-language instructions with coherent, semantically grounded, and physically stable behaviors across multiple embodiments. These results position \texttt{FRoM-W1} as a step toward practical general-purpose humanoid intelligence.

In addition to demonstrating the overall effectiveness of our framework, we have also identified and highlighted several shortcomings within the field that need further improvement. These include the need for larger and higher-quality human motion datasets to enable models to generate actions that are more aligned with textual descriptions, produce motions more suitable for real-world deployment, and ensure that general motion trackers can better follow the noisy motions generated by models.
These all require our entire community to work together for further improvement.

% \section*{Acknowledgements}
% We extend our gratitude to Biao Jiang for discussions and assistance regarding the motion generation model section, to Tairan He and Ziwen Zhuang for their discussions and help in the motion tracking model section, to Zhaoye Fei for assistance and support in computing resources, and to Tao Ji for discussions and help.

%%%%%%%%%%%%%%%%%%%%%%%%%%%%%%%%%%%%%%%%%%%%%%%%%%%%%%%%%%%%
% \bibliographystyle{rusnat}
\bibliographystyle{unsrtnat} %plain， plainnat
\bibliography{main}

\clearpage
% \newpage
\appendix

\section*{\centering \LARGE{Appendix}}

\startcontents[app]
\begingroup
\setcounter{tocdepth}{2}
  % \renewcommand{\contentsname}{Appendix Contents}
  % \section*{\contentsname}
  \printcontents[app]{}{1}{}
\endgroup

\section{More Implementation Details of \texttt{H-GPT}}
\label{app:hgpt}
% \subsection{Benchmark and Evaluation Metric Details}
% % \label{app:benchmark_metric}
% \subsection{Baseline Re-implementation Details}
% \label{app:baselines}

\subsection{Training Data Construction for \texttt{H-GPT}}
\label{app:hgpt_dataset}
We use GPT-4o to generate fine-grained CoT data based on language instructions from HumanML3D-X and rendered videos. The model’s prompt is as follows:
\begin{center}
\framebox{
\begin{minipage}{0.97\textwidth}
% \fontsize{10pt}{12pt}\selectfont
\noindent \texttt{Based on the given motion video, make a plan in the first person within 1 to 2 sentences of how to accomplish the motion in the video. You can refer to the motion description [Two people are engaged in a casual chat.] to refine your plan, but the plan should accurately describe the motion in the video.
Here are some examples:
Example 1:
Motion description: A person is waving goodbye.
Plan:
I stand upright with my hands at my sides, then raise my right hand, wave it up and down, and lower it back to its original position.
Example 2:
Motion description: You are an elephant.
Plan:
I take large steps forward, alternating feet, while swinging my hands and swing my head from side to side.
Example 3:
Motion description: a person walks and trips towards his left and then resumes walking.
Plan:
I walk forward, trip to my left, and awkwardly step with my left foot to regain balance. After stabilizing, I adjust my posture and continue walking with my arms swinging on my side.
!!!Only output the plan.!!!
!!!Do not generate uncertainty descriptions in the plan, e.g. '...or...'!!!
!!!Your plan must accurately and concisely describe the given motion video.!!!}
\end{minipage}
}
\label{app:data_format}
\end{center}

We represent the original hand-inclusive whole-body motion poses in the following format:
\begin{itemize}
    \item $\dot{r}^a \in \mathbb{R}$ is root angular velocity along Y-axis; 
    \item $(\dot{r}^x, \dot{r}^z \in \mathbb{R})$ are root linear velocities on XZ-plane;
    \item $r^y \in \mathbb{R}$ is root height; 
    \item $\mathbf{j}^p \in \mathbb{R}^{3j}$, $\mathbf{j}^v \in \mathbb{R}^{3j}$, and $\mathbf{j}^r \in \mathbb{R}^{6j}$ are the local joints' positions, velocities, and rotations in root space, with $j$ denoting the number of joints;
    \item $\mathbf{c}^f \in \mathbb{R}^4$ is binary features obtained by thresholding the heel and toe joint velocities to emphasize the foot ground contacts.
\end{itemize}

\subsection{More Chain-of-Thought Evaluation Cases}
\label{app:hgpt_cot_eval}

Here are more examples of complex instructions:
\begin{center}
\framebox{
\begin{minipage}{0.97\textwidth}
% \fontsize{10pt}{12pt}\selectfont
\noindent 
A person is running in place, then starts to run right and left, while jumping up and down. \\
A person is kneeling down, coiling something in their hands, then stands, pours out of a bottle, and backs away to the right. \\
A person walks forward, stops, and stands drinking something. \\
A person uses their left hand and arm to move to the side, then front ways, and performs other movements, including waving. \\
A person initially stands still, steps backward with both feet, and then slowly walks forward.
\end{minipage}
}
\end{center}

And here are more examples of abstract instructions:
\begin{center}
\framebox{
\begin{minipage}{0.97\textwidth}
\noindent 
Pantomime pulling a heavy rope hand over hand, showing the strain in your arms and back. \\
Show dejection: slump your shoulders, hang your head, and shuffle your feet slowly. \\
Beckon me to come closer with a curling motion of your index finger. \\
Mime picking up a small, delicate cube from the table using a precise pinch grip. \\
Walk to the nearest wall and gently press the palm of your hand against it.
\end{minipage}
}
\end{center}

% \subsection{\texttt{H-GPT} Training Details on Motion-X Dataset}
% \label{app:hgpt_motionx}

% We utilized Motion-X~\cite{DBLP:conf/nips/LinZLCZWZ23} as our base dataset in practice, which contains $\sim$65,000 samples for training and 4,000 samples for validation. We generated CoT based on the motion $\mathcal{M}_h$ and instruction $\mathcal{I}$ of each sample. 

\section{More Implementation Details of \texttt{H-ACT}}
\label{app:hact}

\subsection{Hardware and Deployment Setup}
\label{app:hact_real_setup}

Our real-world deployment consists of two primary components: a high-performance inference workstation responsible for \texttt{H-GPT} model execution, and a full-size humanoid robot with hands used for \texttt{H-ACT} motion execution. We detail both components below.

% model inference
\paragraph{Inference Server.}

All model inference is performed on a high-performance workstation equipped with dual NVIDIA RTX 5090 GPUs with 32GB VRAM each, an AMD Threadripper 7970X CPU, and 128GB of RAM. 
% The inference pipeline is implemented using PyTorch with CUDA backend and achieves real-time performance, with an average per-frame latency of (\textit{X ms}) at (\textit{Y FPS})\hans{need modification}. 
The workstation communicates wirelessly with the onboard compute units of the deployed robots.

\paragraph{Humanoid Robot Platform.}
% support for multiple robots
To demonstrate the generality and practicality of our approach, we deploy our system on two humanoid robot platforms: the Unitree H1 and Unitree G1. Importantly, both deployments require no hardware modification, relying solely on software integration with the robot's existing control and sensing infrastructure.

% h1
\textit{Unitree H1.}
The H1 is a full-size bipedal humanoid robot equipped with 19 degrees of freedom in the body and an additional 2-DOF wrist. The manipulator is Inspire-Robotics 6-DOF dexterous hands, actuated by a DM-J4310-2EC motor at the wrist. For onboard computation, the robot is outfitted with Unitree's official PC2 module, comprising an Intel i7-1355U CPU and 32GB of RAM. The PC2 executes the control policy and handles wireless communication with the external inference server. Additionally, a ZED Mini stereo camera is mounted on the torso for real-time odometry. The camera is connected to the Unitree Orin NX expansion dock, leveraging the ZED SDK's \texttt{positional\_tracking} service to estimate the robot's global pose. The Orin NX communicates with the PC2 through the robot's internal Ethernet switch, enabling low latency and tight integration between perception and control.

% g1
\textit{Unitree G1.}
The G1 platform, in its EDU+ configuration, features 29 degrees of freedom and Unitree 7-DOF three-fingered Dex3-1 hands. For stability and consistent deployment, we follow the manufacturer's manual to lock the waist pitch and roll joints. The control policy runs on the built-in PC2 unit (Jetson Orin NX 16GB), and no external compute is required. Global localization is supported by two interchangeable options: (1) replicating the ZED Mini configuration used on H1, or (2) utilizing the odometry service provided by the Unitree SDK.

% gr1

% overall
% Across both platforms, our deployment strategy is designed to be plug-and-play: \texttt{H-GPT} and \texttt{H-ACT} can interface with the existing humanoid hardware without any modification to firmware or mechanical components. This highlights the system's compatibility with commercial-grade humanoid hardware and its suitability for real-world deployment.

\begin{figure}[t]
    \centering
    \includegraphics[width=0.98\textwidth,page=1,trim=10 10 10 10,clip]{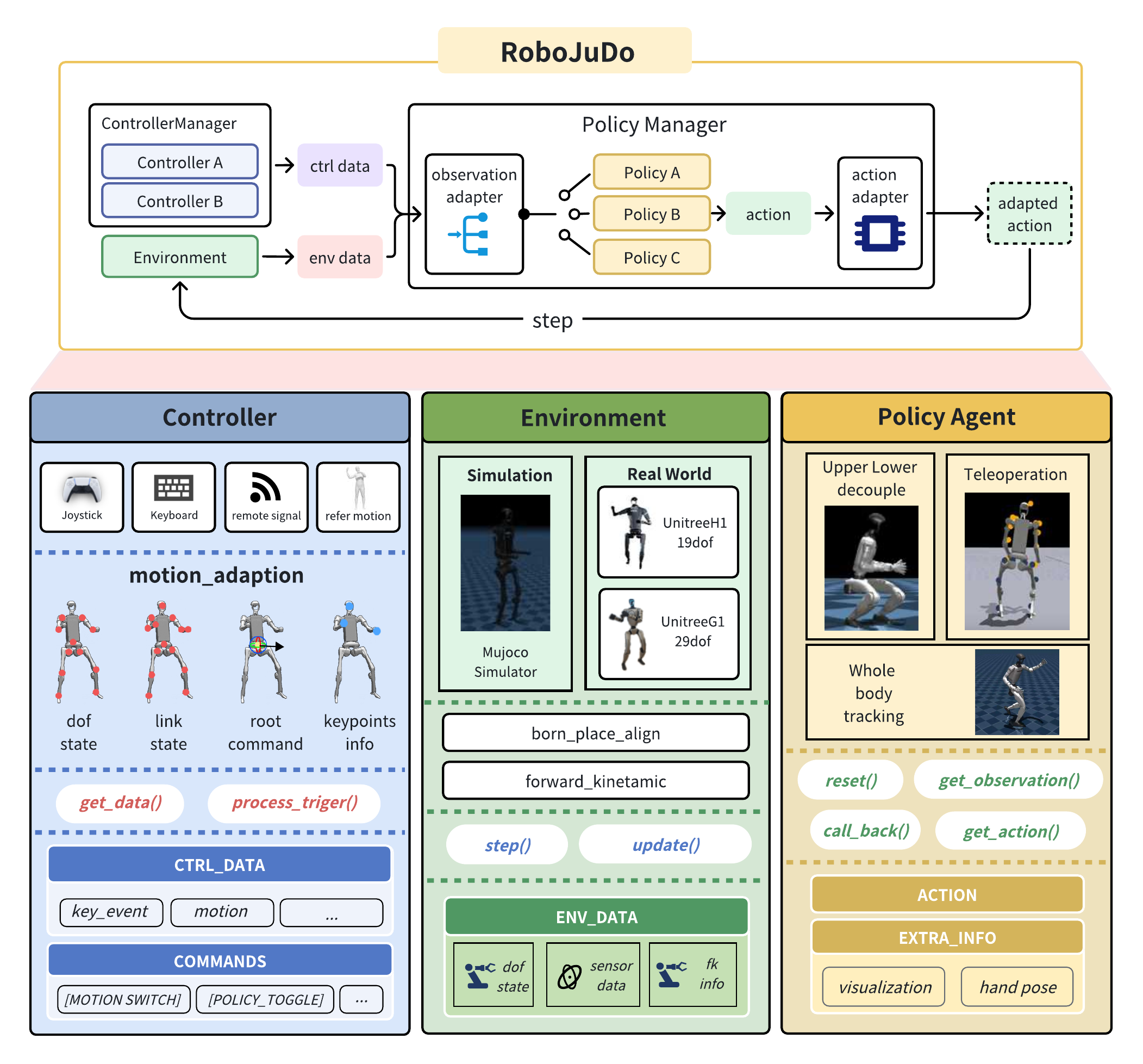}
    \caption{
    An illustration of the \texttt{RoboJudo} framework, including the pipeline control loop and the overall module architecture.
    }
    \label{fig:RoboJuDo}
\end{figure}

\subsection{RoboJuDo Deployment Framework}
\label{app:hact_robojudo}

As illustrated in~\ref{fig:RoboJuDo},\texttt{RoboJudo} adopts a unified module abstraction for \textbf{$C$}, \textbf{$E$}, and \textbf{$\Pi_\theta$}, enabling modular composition and seamless integration across simulation and real-robot platforms.

The \textbf{Controller $C$} standardizes diverse input sources—such as discrete key commands, continuous motion streams, and external triggers—into a consistent command interface.
A hierarchical \textbf{Controller Manager} supports compositional logic, allowing for multi-level triggering (e.g., key combinations) and real-time motion adaption.

The \textbf{Environment $E$} provides a platform-agnostic interface that abstracts robot communication, initialization alignment (e.g., born-place matching), and forward kinematics.
This design ensures functional parity between simulation and real-robot execution, allowing policies to be transferred directly with minimal engineering overhead.

The \textbf{Policy $\Pi_\theta$} defines a unified interface for diverse control strategies. Each policy interacts with both the controller and the environment through standardized APIs — $s_t$ and $m_t$, enabling consistent data flow and behavior representation across heterogeneous implementations.
A dedicated \textbf{Policy Manager} governs multi-policy coordination and adaptation: it automatically aligns each policy’s degree-of-freedom configuration (\textit{DoFConfig}) with the target robot environment, dynamically handling joint remapping and scaling. This mechanism enables seamless switching and blending between heterogeneous policies within a single runtime pipeline, facilitating efficient experimentation and flexible behavior composition.

\paragraph{Controller $C$ module.}
The Controller module unifies multiple input sources, including joysticks, keyboards, remote signals, and reference motion streams from \texttt{H-GPT}.
Raw inputs are collected into a normalized \texttt{CTRL\_DATA} structure (e.g., axes, buttons, motion sequence).
A motion-adaption layer then converts these signals into controller-specific motion descriptors, such as keypoint trajectories, joint states, body-link states, or root locomotion commands.
The module also detects discrete \texttt{COMMANDS} (e.g., motion switching, mode toggles) via \texttt{process\_trigger()}.
This design allows the same downstream pipeline to operate with different human interfaces or autonomous sources by only changing controller configuration, while keeping the motion command format consistent.

\paragraph{Environment $E$ module.}
The Environment module provides a unified interface for both simulated and real humanoids.
On the simulation side, it interfaces with MuJoCo to step the dynamics, while on the real side it communicates with platforms such as Unitree H1 (19~DoF) and Unitree G1 (29~DoF and 23~DoF).
A \texttt{born\_place\_align} utility aligns the initial robot state with the reference motion, and a forward-kinematics service exposes link and keypoint information to other modules.
The core API consists of \texttt{step()} and \texttt{update()}, which advance the simulation or read sensors, and populate \texttt{ENV\_DATA} (including DoF state, sensor measurements, and Forward Kinematics information).
Robot-specific details (DoF layouts, joint limits, PD gains, communication backends) are encapsulated in environment configurations, thus providing simple interface for robot interaction.

\paragraph{Policy Agent $\Pi_\theta$ module.}
The Policy Agent module wraps different whole-body control strategies—upper–lower decoupled control, teleoperation-based control, and full-body motion tracking—behind a common interface.
Each policy exposes standard methods such as \texttt{reset()}, \texttt{get\_observation()}, \texttt{get\_action()}, and optional \texttt{call\_back()} hooks.
The module outputs \texttt{ACTION} (low-level commands) and optionally \texttt{EXTRA\_INFO} (e.g., hand pose, debug information) to facilitate analysis and visualization.

\paragraph{Automatic adaptation via DoFConfig and PolicyWrapper.}
To support multiple robot embodiments without rewriting controllers, \texttt{RoboJuDo} introduces DoFConfig, including joint ordering, PD gains, that unifies interaction of different controllers and robots.
The PolicyWrapper uses this configuration to automatically remap policy actions and observations between the controller’s internal convention and the target robot’s embodiment.
As a result, the same policy implementation can be reused across robots that share similar semantics but differ in DoF layout, with adaptation handled declaratively through configuration rather than manual code changes.

\paragraph{Multi-policy switching and composition.}
\texttt{RoboJuDo} includes a lightweight policy manager that supports runtime switching and composition of multiple policies.
Based on \texttt{COMMANDS} from the Controller module, the manager can route \texttt{ENV\_DATA} and \texttt{CTRL\_DATA} to different Policy Agents, blend their outputs, or trigger hierarchical behaviors.
Because all policies conform to the same $(s_t, m_t) \mapsto a_{t+1}$ abstraction, switching between them does not require changes to the Environment or Controller modules.
This makes it straightforward to benchmark different WBC paradigms on the same motion sequences, or to compose complex behaviors from simpler skills.

Overall, \texttt{RoboJuDo} provides a modular, configuration-driven pipeline that connects diverse input devices, simulation/real environments, and heterogeneous control policies.
Combined with \texttt{H-GPT} and \texttt{H-ACT}, it forms a unified execution stack that can route generated motions through various controllers and embodiments with minimal engineering effort.

\subsection{Sim2Real Transfer Details}
\label{app:hact_real}

During simulation-based training, we observed practical techniques
% ("tricks")
that could significantly improve the performance and transferability of policies to real-world robotic platforms.

\paragraph{Handling Low-Inertia Joints}
One key issue arises from low-inertia joints, such as those in the wrists or head. In simulation, these joints often exhibit high-frequency oscillations, limited by the numerical accuracy of the physics engine's rigid body solver. These oscillations can destabilize training, hinder policy convergence, and exacerbate the gap between simulation and reality.

To mitigate this, we recommend removing or freezing certain low-inertia degrees of freedom during training. For instance, wrist and head joints can be excluded from learning and reintroduced during real-world deployment with separate low-level controllers or scripted behaviors. This strategy has proven effective in improving the overall stability and robustness of the learned policy.

\end{document}